\documentclass[11pt,a4paper]{article}

\usepackage{geometry}
\usepackage{microtype}
\usepackage{breakurl}
\usepackage{graphicx}
\usepackage{epstopdf}
\usepackage{subcaption}
\usepackage{hyperref}
\usepackage{amsfonts}
\usepackage{amsthm}
\usepackage{amsmath}
\usepackage{amssymb}
\usepackage[square,sort,comma,numbers]{natbib}
\usepackage{authblk}
\usepackage{indentfirst}
\usepackage{appendix}

\newcommand{\mm}{{\mathcal{M}}}
\newcommand{\hh}{{\mathcal{H}}}
\newcommand{\mse}{\text{MSE}}

\newcommand{\cov}{\text{Cov}}
\newcommand{\wtid}{\widetilde}
\newcommand{\txm}{T_p\mathcal{M}}

\newcommand{\hpca}{h_{\text{PCA}}}
\newcommand{\bxi}{\mathbf{\Xi}}
\newcommand{\bx}{\mathbf{\Delta}}
\newcommand{\bk}{\mathbf{K}}
\newcommand{\bl}{\mathbf{L}}
\newcommand{\bd}{\mathbf{D}}
\newcommand{\br}{\mathbf{r}}
\newcommand{\bu}{\mathbf{u}}
\newcommand{\bv}{\mathbf{v}}
\newcommand{\bw}{\mathbf{w}}
\newcommand{\bz}{\mathbf{z}}
\newcommand{\bt}{\mathbf{\Theta}}
\newcommand{\what}{\widehat}
\newcommand{\ef}{\mathbb{E}_f}
\newcommand{\tr}{\text{Tr}}
\newcommand{\diag}{\text{diag}}
\newcommand{\rt}{\textrm{t}}

\title{Efficient Weingarten Map and Curvature Estimation on Manifolds}
\author[1]{Yueqi Cao }
\author[3,4]{Didong Li }
\author[1,2]{Huafei Sun}
\author[5]{Amir H Assadi}
\author[1]{Shiqiang Zhang}

\affil[1]{School of Mathematics and Statistics, Beijing Institute of technology}
\affil[2]{Beijing Key Laboratory on MCAACI}
\affil[3]{Department of Computer Science, Princeton University}
\affil[4]{Department of Biostatistics, University of California, Los Angeles}
\affil[5]{Department of Mathematics, University of Wisconsin Madison}

\date{}

\begin{document}

\maketitle

\numberwithin{equation}{section}
  \newtheorem{theorem}{Theorem}
  \newtheorem{prop}{Proposition}
  \newtheorem{lemma}{Lemma}
  \newtheorem{example}{Example}
  \newtheorem{coro}{Corollary}
  \newtheorem{remark}{Remark}
  \newtheorem{definition}{Definition}

\begin{abstract}
In this paper, we propose an efficient method to estimate the Weingarten map for point cloud data sampled from manifold embedded in Euclidean space. A statistical model is established to analyze the asymptotic property of the estimator. In particular, we show the convergence rate as the sample size tends to infinity.  We verify the convergence rate through simulated data and apply the estimated Weingarten map to curvature estimation and point cloud simplification to multiple real data sets. 
\end{abstract}

\section{Introduction}
\label{introduction}

In the literature of manifold learning, data in high dimensional Euclidean space are always assumed to lie on a low dimensional manifold $\mm$. The inference of geometry and topology of $\mm$ becomes critical in the understanding and application of the data such as dimension reduction, clustering and visualization. Topological quantities often involve persistent homology \cite{edelsbrunner2008persistent,zomorodian2005computing}, homotopy groups \cite{letscher2012persistent} and fundamental groups \cite{batan2019van}, while geometric quantities include intrinsic dimension \cite{little2017multiscale}, tangent space \cite{Wu2012connection}, geodesics \cite{li2019geodesic}, Laplacian operators \cite{belkin2004semi,belkin2007convergence} and curvature \cite{aamari2019nonasymptotic,buet2018discretization}. There are also hot topics on the estimation of volume and support. This paper focuses on the estimation of the Weingarten map, or the second fundamental form, for point clouds sampled from submanifolds embedded in Euclidean spaces. The second fundamental form is a useful tool to study Riemannian manifold extrinsically. On the one hand, since there is no prior information about the manifold, it prohibits the intrinsic way to study the manifold from sampling data. On the other hand, the second fundamental form is closely related to other geometric quantities of the manifold. For example, in \cite{niyogi2008finding} the authors showed that the operator norm controls the reach of the manifold. In \cite{absil2013extrinsic} they proposed a simple way to compute the Riemannian hessian based on the Weingarten map. Furthermore, it is well known that the second fundamental form measures how manifolds curve in the ambient space. Once the second fundamental form is known, we can compute all kinds of curvature from it. 

Our motivation comes from the need to estimate the curvature for unstructured point cloud data. Efficient estimation of (Gaussian/mean/principal) curvature for point cloud data is an important but difficult problem. In geometry, curvature contains much information of the underlying space of an unordered point set. Therefore it provides prior information in many  applications such as surface segmentation \cite{woo2002new,rabbani2006segmentation}, surface reconstruction \cite{wang2006fitting,Berger2016A,TANG201828}, shape inference \cite{Curvatureaugmentedtensorvoting,Dey2002Shape}, point cloud simplification \cite{discretecurvaturenorm,Efficientsimplification} and feature extraction \cite{Automaticdatasegmentation}. However, methods used to estimate curvature are restricted. Direct computation on point clouds often requires local parametrization. One needs to fit a local parametric surface first. Then the curvature is obtained by substituting the coefficients into an analytical formula.  Another way is to estimate curvature after surface reconstruction which turns point clouds into triangular meshes or level sets of some distance function \cite{Inferringsurfacetraceanddifferentialstructure,Hoppe,Levin1998The,OnsurfacenormalandGaussiancurvature}.
However, there is little theory about estimation error analysis like convergence rate. This is not surprising since these existing algorithms are not aiming at minimizing the estimation error of the curvature, but instead to minimize the error of the surface approximation. In addition, even when surfaces are close to each other under the Euclidean distance, their curvature might not be close to each other. For example, we can perturb a straight line a little bit so that the curvature is far away from zero. As a result, a more direct and efficient approach with theoretical guarantee on the estimation error is needed.

Recently several methods have been proposed to estimate the second fundamental form. In \cite{buet2018discretization}, the authors proposed the notions of generalized second fundamental form and mean curvature based on the theory of varifolds. The generalized mean curvature descends to the classical mean curvature if the varifold is the standard measure on a submanifold. Under specific conditions, they proved that the optimal convergence rate is $O(n^{-1/2})$ where $n$ is the sample size, regardless of the dimension of the varifold. In \cite{aamari2019nonasymptotic}, the authors used polynomial fitting to estimate the tangent space, the second fundamental form and the support of manifold simultaneously. Under some assumptions for the regularity of the manifold and sampling, they proved the convergence rate for the second fundamental form is $O(n^{-(k-2)/m})$ where $k$ is regularity parameter and $m$ is the dimension of manifold. In computer vision, several approaches are based on triangular meshes rather than point clouds. For example, S. Rusinkiewicz approximates the second fundamental form per triangle for meshes \cite{rusinkiewicz2004estimating}. J. Berkmann and T. Caelli proposed two covariance matrices to approximate the second fundamental form \cite{berkmann1994computation}. 
For curvature estimation, there are methods proposed without the need of estimation of second fundamental form. In \cite{merigot2011voronoi}, the authors introduced the definition of Voronoi covariance measure (VCM) to estimate the curvature for noisy point clouds.
  Though a stability theorem is proved, there is no assertion on convergence rate. In the classic paper \cite{taubin1995estimating}, G. Taubin defined a matrix by an integration formula.
 As illustrated in \cite{lange2005anisotropic}, this is nothing but the Weingarten map. This formula is adopted by \cite{lange2005anisotropic} to estimate the principal curvature on point sets. The authors then proposed a method for anisotropic fairing of a point sampled surface using mean curvature flow.

\subsection{Our Contribution}

In this paper, we proposed a two-step procedure to estimate the Weingarten map for point clouds sampled from submanifolds embedded in Euclidean spaces. Firstly, we use local PCA to estimate the tangent spaces and normal spaces. Secondly, we apply a least-square method to compute the matrix representing the Weingarten map under the estimated tangent basis. The algorithm is general for point clouds in any dimension, and is efficient to implement due to low complexity. 

A statistical model is set up to analyze the convergence rate for the Weingarten map estimator. Under the assumption of exact tangent spaces and given normal vector field, we proved that if the bandwidth is chosen to be $O(n^{-1/(m+4)})$, then the optimal convergence rate will be $O(n^{-4/(m+4)})$. Other than kernel method, we also discussed the $k$-nearest-neighbor method, which is more convenient to use in practice. Compared with the method proposed in \cite{buet2018discretization}, our method converges faster in low dimension. In comparison with the estimator proposed in \cite{aamari2019nonasymptotic}, our method gives a closed form and easier to compute. 

The convergence rate is verified by numerical experiments on two synthetic data sets. We also compare WME with the traditional quadratic fitting method, the state-of-art algorithm in this literature. Our method yields better results than  the quadratic fitting in both MSE and robustness. As an application, we propose a curvature-based clustering method in point cloud simplification. Furthermore, we reconstruct surfaces based on the simplified point clouds to give a visible comparison. Three real data sets are tested to show the gain of our WME algorithm.

\paragraph{Outlines}This paper is organized as follows. We introduce our WME method in section \ref{method} followed by a statistical model to analyze the convergence rate in section \ref{error estimation}. In section \ref{numericalexperiments}, we verify the convergence rate and compare our method with quadratic surface fitting method using synthetic data. Applications to brain cortical data and experiments on point cloud simplification are given in section \ref{simplification}. In section \ref{discussion}, we discuss the possible applications of WME algorithm in future works.


\section{Algorithm Description}
\label{method}

Let $\mm\subseteq \mathbb{E}^d$ be an $m$-dimensional submanifold in a $d$-dimensional Euclidean space with induced Riemannian metric. At each point $p$ we have the following decomposition
\begin{equation}
	T_p\mm\oplus T_p^\perp \mm=\mathbb{E}^d
\end{equation}
Let $\overline{\nabla}$ be the standard connection in $\mathbb{E}^d$. For any normal vector $\xi\in T_p^\perp \mm$, extend $\xi$ to be a normal vector field $\tilde{\xi}$ on $\mm$. \emph{The Weingarten map} or \emph{the shape operator} at $p$ with respect to $\xi$ is defined as
\begin{equation}
	\begin{aligned}
	&A_\xi:T_p\mm\to T_p\mm\\
	&A_\xi(X)=-(\overline{\nabla}_X\tilde{\xi})^\top
	\end{aligned}
\end{equation}
where $^\top:E^d\to T_p\mm$ is the orthogonal projection to the tangent space. It can be verified that the definition of $A_\xi$ is independent of the extension of $\xi$ (see appendix \ref{appen-weingarten}). 

\begin{example}
Let $\hh^{d-1}\subseteq \mathbb{E}^d$ be a hypersurface, and $\tilde{\xi}$ be a unit normal vector field on $\hh^{d-1}$. \emph{The Gauss map} $g:\hh^{d-1}\to \mathbb{S}^{d-1}$ sending any point on the hypersurface to a point on the unit sphere is defined by $g(p)=\tilde{\xi}_p$. For any $X\in T_p\hh^{d-1}$, we have
\begin{equation}
	A_{\tilde{\xi}_p}(X)=-{\rm d}g(X)
\end{equation}
that is, $-A_{\tilde{\xi}_p}$ is the tangent map of the Gauss map.
\end{example}

The Weingarten map measures the variation of the normal vector field. In fact, from the following `Taylor expansion' of a normal vector field, we can see that the Weingarten map plays the role similar to the derivative of a function.

\begin{prop}\label{prop}
	 Let $\tilde{\xi}$ be a normal vector field on the submanifold $\mm\subseteq\mathbb{R}^d$. Suppose that $p$ is a point on $\mm$ and $q$ is any point within the geodesic neighborhood of $p$. Denote $^\top$ to be the orthogonal projection to the tangent space at $p$. We have  
	\begin{equation}\label{maineq}
	(\tilde{\xi}_q-\tilde{\xi}_p)^\top=-A_{\tilde{\xi}_p}\big((q-p)^\top\big)+O(\|q-p\|^2)
	\end{equation}
\end{prop}

\begin{proof}
	Let $\mathbf{r}:U^m\subseteq \mathbb{R}^m\to \mm$ be the exponential map such that $\mathbf{r}(0)=p$. Denote $\mathbf{u}=(u^1,\cdots,u^m)\in U^m$. 
	Then the vector fields $\{\frac{\partial\mathbf{r}}{\partial u^i}\}_{i=1}^m$ form a local tangent frame. Denote $\tilde{\xi}(\mathbf{r}(\mathbf{u}))$ by $\tilde{\xi}(\mathbf{u})$. Consider the following expansions
	\begin{equation}\label{normalexpansion}
	\begin{aligned}
	&\tilde{\xi}(\mathbf{u})=\tilde{\xi}(0)+\sum_{i=1}^{m}u^i\frac{\partial\tilde{\xi}}{\partial u^i}(0)+O(\|\mathbf{u}\|^2)\\
	&\mathbf{r}(\mathbf{u})=\mathbf{r}(0)+\sum_{i=1}^{m}u^i\frac{\partial\mathbf{r}}{\partial u^i}(0)+O(\|\mathbf{u}\|^2)
	\end{aligned}
	\end{equation}
	By definition of $A_{\tilde{\xi}_p}$, we have
	\begin{equation}
		A_{\tilde{\xi}_p}\big(\frac{\partial\mathbf{r}}{\partial u^i}(0)\big)=-\big(\overline{\nabla}_{\frac{\partial\mathbf{r}}{\partial u^i}(0)}\tilde{\xi}\big)^\top=-\big(\frac{\partial\tilde{\xi}}{\partial u^i}(0)\big)^\top
	\end{equation} 
	 Substituting to \eqref{normalexpansion}, we obtain that
	\begin{equation}
	\begin{aligned}
	\big(\tilde{\xi}(\mathbf{u})-\tilde{\xi}(0)\big)^\top=&\big(\sum_{i=1}^{m}u^i\frac{\partial\tilde{\xi}}{\partial u^i}(0)+O(\|\mathbf{u}\|^2)\big)^\top
	=-A_{\tilde{\xi}_p}\big((\mathbf{r}(\mathbf{u})-\mathbf{r}(0))^\top\big)+O(\|\mathbf{u}\|^2)\\
	\end{aligned}
	\end{equation}
	Note that $\|\mathbf{u}\|$ represents the geodesic distance on the manifold by the property of the exponential map. According to the proposition \ref{exponent-map-prop} in appendix \ref{appendix-exponetial}, the geodesic distance can be approximated by the Euclidean distance in the same order. Therefore, \eqref{maineq} follows.
\end{proof}

 Assume that $n$ points $x_1,\cdots, x_n$ viewed as points in $\mathbb{R}^d$ are independently sampled from some distribution on $\mm$. The object is to estimate the Weingarten maps at each $x_i$ ($i=1,2,\cdots,n$). Thus, firstly, we need to estimate the tangent and normal spaces. Otherwise, the estimation of Weingarten maps does not make sense without specifying the normal directions. Local PCA is an extensively used method in manifold learning to estimate tangent and normal spaces, whose effectiveness and consistency is well understood (see \cite{Wu2012connection}). Secondly, after tangent space estimation, proposition \ref{prop} indicates a simple linear model to estimate the Weingarten maps from the data points, which can be resolved by a least-square method. Therefore, our WME algorithm is a two-step procedure where each step can be proceeded efficiently.
 
 \textbf{Local PCA}: For every data point $x_i$ ($i=1,2,\cdots,n$) we try to estimate a basis $e_i^1,\cdots,e_i^m$ to the tangent space and a basis $\xi_i^1,\cdots,\xi_i^{d-m}$ to the normal space. Fix a parameter $\hpca>0$ and define the following set$I_i=\{j\in\mathbb{N}|\|x_j-x_i\|\le \hpca\}$.
 The local covariance matrix is defined as 
 \begin{equation}
 	\cov = \sum_{j\in I_i} (x_j-\bar{x}_i)(x_j-\bar{x}_i)^{\rm t}
 \end{equation} 
 where each data point is regarded as a column vector in $\mathbb{R}^d$ and $\bar{x}_i=\frac{1}{|I_i|}\sum_{j\in I_i}x_j$ is the mean of neighboring points. The first $m$ eigenvectors of $\cov$ corresponding to the first $m$ largest eigenvalues constitute the basis to the tangent space at $x_i$, while the last $d-m$ eigenvectors corresponding to the last $d-m$ smallest eigenvalues form the basis to the normal space at $x_i$.
 
 \textbf{Normal Vector Extension}: Suppose that we pick up the normal vector $\xi_i^\alpha$ ($i=1,2,\cdots, m$, $\alpha=1,2,\cdots,d-m$) and want to estimate the Weingarten map with respect to $\xi_i^\alpha$. Extend $\xi_i^\alpha$ to be a normal vector field $\tilde{\xi}^\alpha$ on $\mm$ by setting
 \begin{equation}
 	\tilde{\xi}^\alpha_{x_j}=\sum_{\beta=1}^{d-m}\langle \xi_i^\alpha,\xi_j^\beta \rangle\xi_j^\beta
 \end{equation}
for $j=1,2,\cdots,n$. That is, we project $\xi_i^\alpha$ to the normal space at $x_j$. Since projection varies smoothly with respect to the points on the manifold, it results in a smooth normal vector field.
 
Assume $K:\mathbb{R}\to \mathbb{R}$ is a twice differentiable function supported on $[0,1]$. For example, the truncated Gaussian kernel on $[0,1]$ or Epanechnikov kernel. 
Let $y\in \mathbb{R}^d$. Given $h>0$, define $K_h:\mathbb{R}^d\to \mathbb{R}$ to be 
\begin{equation}
K_h(y)=\frac{1}{h^m}K(\frac{\|y\|}{h})
\end{equation}
Let $E_i=[e_i^1,\cdots,e_i^m]$ be the matrix consisting of the basis to the tangent space at $x_i$. According to proposition \ref{prop}, if $x_j$ is close to $x_i$,
  \begin{equation}\label{opfundamental}
    (\tilde{\xi}^\alpha_{x_j}-\xi_i^\alpha)^{\rm t}E_i= -(x_j-x_i)^{\rm t}E_iA_{\xi_i^\alpha}+O(\|x_j-x_i\|^2)
  \end{equation}
  where $A_{\xi_i^\alpha}$ is understood as $m\times m$ matrix. Therefore, we want to find a matrix $\wtid{A}_{\xi_i^\alpha}$ which minimizes the following residual
\begin{equation}\label{optimization}
	\sum_{j=1}^{n}\|(\tilde{\xi}^\alpha_{x_j}+\xi_i^\alpha)^{\rm t}E_i-(x_j-x_i)^{\rm t}E_i\wtid{A}_{\xi_i^\alpha}\|^2K_h(x_j-x_i)
\end{equation}
Set
\begin{equation}
  \begin{aligned}
  	&\wtid{\Delta}_i=E_i^{\rm t}\left[x_1-x_i,\cdots,x_n-x_i\right]\\
    &\wtid{\Xi}_i^\alpha=E_i^{\rm t}\left[\tilde{\xi}^\alpha_{x_1}-\xi_i^\alpha,\cdots,\tilde{\xi}^\alpha_{x_n}-\xi_i^\alpha\right]\\
    &W_i = \diag\{K_h(x_1-x_i),\cdots,K_h(x_n-x_i)\}
  \end{aligned}
\end{equation}
Then the solution of \eqref{optimization} is given in the following closed form
\begin{equation}\label{wtidA}
	\wtid{A}_{\xi_i^\alpha}=-\wtid{\Xi}_i^\alpha W_i\wtid{\Delta}_i^{\rm t}(\wtid{\Delta}_i W_i\wtid{\Delta}_i^{\rm t})^{-1}
\end{equation}

\begin{remark}
 The above method gives the estimation of the Weingarten maps at each point with respect to all normal basis. Since each Weingarten map is an $m\times m$ matrix and the normal basis consists of $d-m$ vectors, the whole procedure gives $m\times m\times (d-m)$ coefficients (in fact $\frac{m(m+1)}{2}\times (d-m)$ coefficients since the Weingarten map is symmetric) at each point. Using these coefficients we can give the estimation of second fundamental form and mean curvature and sectional curvature at each point. If we assume the underlying manifold is of low dimension, then all the coefficients are about the size $O(dn)$ where $d$ is the dimension of the ambient space and $n$ is the size of the point cloud.
\end{remark}

\begin{remark}
	 Since the Weingarten map is a self-adjoint operator on the tangent space, the matrix $\wtid{A}_{\xi_i^\alpha}$ should be symmetric. It is natural to solve \eqref{optimization} on the space of symmetric matrices. However, as we will prove later, the solution of \eqref{optimization} converges to the true matrix. It is not necessary to solve a more complex optimization problem on the space of symmetric matrices. In cases where symmetry is important we can always use the symmetrization $\frac{1}{2}(\wtid{A}_{\xi_i^\alpha}^{\rm t}+\wtid{A}_{\xi_i^\alpha})$.
\end{remark}

\section{Convergence Rate}\label{error estimation}

\subsection{Statistical Modeling}

Let $\xi$ be a normal vector field on $\mm$, and $P$ be a random vector valued in $\mm$ with smooth positive density function $f$. Fix a point $p\in\mm$.
To avoid complexity on notations, we drop the subscripts and  $A:\txm\to\txm$ is always understood as the Weingarten map (or its matrix representation if a basis is specified) at $p$ with respect to $\xi_p$. We rewrite proposition \ref{prop} as follows: when $P$ is within the normal neighborhood of $p$, we have 
\begin{equation}\label{modelequation}
(\xi_P-\xi_p)^\top=-A\big((P-p)^\top\big)+\eta(P)\|P-p\|^2
\end{equation}
where  $\eta:\mm\to \mathbb{R}^m$ is assumed to be a bounded smooth function in the neighborhood of $p$. In addition,  assume that a basis $e_1,e_2,\cdots,e_m$ for the tangent space $\txm$ is given.
Set $\bxi=(\xi_P-\xi_p)^\top$, $\bx=(P-p)^\top$ to be the vectors representing the coordinates under the basis, and set $\bk_P=K_h(P-p)$. Consider the following optimization problem
\begin{equation}
\mathop{\arg\min}_{\mathbf{A}\in\mathbb{R}^{m\times m}}\ef\left[\|\bxi+\mathbf{A}\bx\|^2\bk_P\right]
\end{equation}
That is, we want to find the minimizer of the function 
\begin{equation}
\begin{aligned}
F(\mathbf{A})&=\ef\left[\tr(\bxi\bxi^{\rm t}+\bxi\bx^{\rm t}\mathbf{A}^{\rm t}+\mathbf{A}\bx\bxi^{\rm t}+\mathbf{A}\bx\bx^{\rm t}\mathbf{A}^{\rm t})\bk_P\right]\\
&=\tr(\ef[\bxi\bxi^{\rm t}\bk_P])+2\tr(\ef[\bxi\bx^{\rm t}\bk_P]\mathbf{A})+\tr(\mathbf{A}\ef[\bx\bx^{\rm t}\bk_P]\mathbf{A}^{\rm t})
\end{aligned}
\end{equation}
By setting $dF/d\mathbf{A}=0$, the population solution can be given in the following closed form
\begin{equation}
\mathbf{A}=-\ef[\bxi\bx^\rt\bk_P](\ef[\bx\bx^\rt\bk_P])^{-1}=-\bl\bd^{-1}
\end{equation}
where we have set $\bl=\ef[\bxi\bx^\rt\bk_P]$ and $\bd=\ef[\bx\bx^\rt\bk_P]$. Denote the coordinate components $\bxi\cdot e_j$ by $(\bxi)_j$ and $\bx\cdot e_j$ by $(\bx)_j$ for $j=1,2,\cdots,m$. In matrix form we have
\begin{equation}
\begin{aligned}
&\bl=\begin{bmatrix}
\ef[(\bxi)_1(\bx)_1\bk_P]&\cdots&\ef[(\bxi)_1(\bx)_m\bk_P]\\
\vdots&\ddots&\vdots\\
\ef[(\bxi)_m(\bx)_1\bk_P]&\cdots&\ef[(\bxi)_m(\bx)_m\bk_P]\\
\end{bmatrix}\\
&\bd=\begin{bmatrix}
\ef[(\bx)_1^2\bk_P]&\cdots&\ef[(\bx)_1(\bx)_m\bk_P]\\
\vdots&\ddots&\vdots\\
\ef[(\bx)_m(\bx)_1\bk_P]&\cdots&\ef[(\bx)_m^2\bk_P]\\
\end{bmatrix}\\
\end{aligned}
\end{equation}
Let $x_1,\cdots, x_n$ be i.i.d. samples from $P$. Let $\Xi_i,\Delta_i$ be the quantities obtained by replacing the random vector in $\bxi,\bx$ with samples for $i=1,2,\cdots,n$. Substituting the expectation by empirical mean, we obtain the empirical solution
\begin{equation}
\wtid{A}=-\wtid{L}\wtid{D}^{-1}
\end{equation} 
where 
\begin{equation}
\begin{aligned}
&\wtid{L}=\begin{bmatrix}
\frac{1}{n}\sum_{i=1}^{n}(\Xi_i)_1(\Delta_i)_1\bk_{x_i}&\cdots&\frac{1}{n}\sum_{i=1}^{n}(\Xi_i)_1(\Delta_i)_m\bk_{x_i}\\
\vdots&\ddots&\vdots\\
\frac{1}{n}\sum_{i=1}^{n}(\Xi_i)_m(\Delta_i)_1\bk_{x_i}&\cdots&\frac{1}{n}\sum_{i=1}^{n}(\Xi_i)_m(\Delta_i)_m\bk_{x_i}\\
\end{bmatrix}\\
&\wtid{D}=\begin{bmatrix}
\frac{1}{n}\sum_{i=1}^{n}(\Delta_i)^2_1\bk_{x_i}&\cdots&\frac{1}{n}\sum_{i=1}^{n}(\Delta_i)_1(\Delta_i)_m\bk_{x_i}\\
\vdots&\ddots&\vdots\\
\frac{1}{n}\sum_{i=1}^{n}(\Delta_i)_m(\Delta_i)_1\bk_{x_i}&\cdots&\frac{1}{n}\sum_{i=1}^{n}(\Delta_i)^2_m\bk_{x_i}\\
\end{bmatrix}\\
\end{aligned}
\end{equation}
It is easy to check that the empirical solution given here is the same as the one given in \eqref{wtidA}. That is, it is the solution of the following empirical optimization problem
\begin{equation}
\mathop{\arg\min}_{\wtid{A}\in\mathbb{R}^{m\times m}}\sum_{i=1}^{n}\|\Xi_i-\wtid{A}\Delta_i\|^2\bk_{x_i}
\end{equation}
Finally, the mean square error (MSE) is defined as 
\begin{equation}\label{msedefinition}
\mse=\ef\left[\|\wtid{A}-A\|_F^2\right]\le 2\bigg(\underbrace{\ef\left[\|\wtid{A}-\mathbf{A}\|_F^2\right]}_{\text{Variance}}+\underbrace{\|A-\mathbf{A}\|_F^2}_{\text{Bias}^2}\bigg)
\end{equation}
where $\|\cdot\|_F$ denotes the Frobenius norm.

\subsection{Mean Square Error}

In the estimation of either variance or bias, the main obstacle is how to control the norm of inverse matrices $\bd^{-1}$ and $\wtid{D}^{-1}$. The method is somewhat similar to that in kernel density estimation. When the bandwidth $h$ is small, the integration is taken over a normal neighborhood near $p$. Using the parametrization of the exponential map, the integral domain is the tangent space and we can use Taylor expansion to find the leading terms. Before that we notice the following fact concerning Euclidean distance and manifold distance, which can be found in \cite{ozakin2009submanifold}.   
\begin{lemma}\label{distance}
	Let $d_p$ and $\rho_p$ be the Riemannian distance and Euclidean distance to $p$, respectively. There exists a function $R_p(h)$ and positive constants $\delta_{R_p}$, $C_{R_p}$ such that when $h<\delta_{R_p}$, $d_p(y)\le R_p(h)$ for all $y$ with $\rho_p(y)\le h$, and furthermore,
	\begin{equation}
	h\le R_p(h)\le h+C_{R_p}h^3
	\end{equation}
\end{lemma}
Lemma \ref{distance} indicates that for $h$ small enough if $\|P-p\|\le h$ then there exists a function such that the geodesic distance between $P$ and $p$ is controlled by $R_p(h)$. First we give the estimation of $\bd^{-1}$. Some properties of exponential map are presented in appendix \ref{appendix-exponetial}. 
\begin{lemma}\label{estimatebd-lemma}
	There exists a positive constant $h_0$ such that when $h<h_0$,
	\begin{equation}\label{bdequalshI}
	\bd=h^2\left(f(p)\int_{\|\bz\|\le 1}(z^1)^2K(\|\bz\|){\rm d}\bz\right)(I+o(1))
	\end{equation}
	where $I$ is the identity matrix. Thus, the inverse is given by
	\begin{equation}
	\bd^{-1}=\frac{1}{h^2\left(f(p)\int_{\|\bz\|\le 1}(z^1)^2K(\|\bz\|){\rm d}\bz\right)}(I+o(1))
	\end{equation}
\end{lemma} 
\begin{proof}
	Pick $h_0$ such that for $h<h_0$ the exponential map is well defined on the geodesic ball of radius $R_p(h)$. For an arbitrary element $\bd_{kl}$, let $\br$ be the exponential map. We have
	\begin{equation}\label{estimateD1}
	\begin{aligned}
	&\ef[(\bx)_k(\bx)_l\bk_P]=\int_{\mathcal{M}}((P-p)\cdot e_k)((P-p)\cdot e_l)K_h(P-p)f(P)dv\\
	&=\frac{1}{h^m}\int_{\|\bu\|\le R_x(h)}(u^k+o(\|\bu\|))(u^l+o(\|\bu\|))K(\frac{\|\bu\|+o(\|\bu\|)}{h})f(\br(\bu))\sqrt{\det(g(\bu))}{\rm d}\bu
	\end{aligned}
	\end{equation} 
	where $g(\bu)$ denotes the Riemannian metric matrix. In normal coordinates, $g_{ij}(\bu)=\delta_{ij}+o(\|\bu\|)$. Changing the variable of integration to $\bu=h\bz$, we obtain that
	\begin{equation}\label{estimateD2}
	\eqref{estimateD1}=\int_{\|\bz\|\le R_x(h)/h}(hz^k+o(h\|\bz\|))(hz^l+o(h\|\bz\|))K(\|\bz\|+o(h\|\bz\|)/h)f(\br(h\bz))\sqrt{g(h\bz)}{\rm d}\bz
	\end{equation}
	The integral domain can be divided into two parts: the first is the unit ball $Q_1=\{\bz|\|\bz\|\le 1\}$ and the second is $Q_2=\{\bz|1\le\|\bz\|\le R_x(h)/h\}$. On $Q_1$, the integration is 
	\begin{equation}
	h^2\left(\int_{\|\bz\|\le 1}z^kz^lK(\|z\|)f(x){\rm d}\bz+o(1)\right)=h^2(\delta_{kl}f(x)\int_{\|\bz\|\le 1 }(z^1)^2K(\|z\|){\rm d}\bz+o(1))
	\end{equation}
	where we have used the symmetry of integration and $\delta_{kl}$ denotes the Kronecker delta. 
	On $Q_2$, note that the integrand is a quantity of $O(h^2)$ whereas the area of integral domain is $O(h^2)$ by lemma \ref{distance}. Overall, we have shown that
	\begin{equation}
	\ef[(\bx)_k(\bx)_l\bk_P]=h^2(\delta_{kl}f(x)\int_{\|\bz\|\le 1 }(z^1)^2K(\|z\|){\rm d}\bz+o(1))
	\end{equation} 
	which proves equation \eqref{bdequalshI}. The inverse matrix is given by the identity $(I-D)^{-1}=\sum_{k=0}^{\infty}D^k$ for small $D$.
\end{proof}

Using the presented results, we are able to give the convergence order of bias.
\begin{lemma}
	As $h\to 0$, $\|\mathbf{A}-A\|_F^2= O(h^4)$.
\end{lemma}
\begin{proof}
	According to the model assumption \eqref{modelequation}, if we set $\bt=\bxi-A\bx=\eta(P)\|P-p\|^2$, then we have
	\begin{equation}
	\mathbf{A}=\ef[(A\bx+\bt)\bx^\rt\bk_P](\ef[\bx\bx^\rt\bk_P])^{-1}=A+\ef[\bt\bx^\rt\bk_P](\ef[\bx\bx^\rt\bk_P])^{-1}
	\end{equation}
	It suffices to estimate $\ef[\bt\bx^\rt\bk_P]$. As in the proof of lemma \ref{estimatebd-lemma}, for $h<h_0$, the integration is 
	\begin{equation}
	\begin{aligned}
	&\ef[(\bt)_k(\bx)_l\bk_P]=\int_{\mathcal{M}}\eta^k(P)\|P-p\|^2((P-p)\cdot e_l)K_h(P-p)f(P)dv\\
	&=\frac{1}{h^m}\int_{\|\bu\|\le R_x(h)}\eta^k(\br(\bu))(\|\bu\|^2+o(\|\bu\|^2))(u^l+O(\|\bu\|^2))K(\frac{\|\bu\|+o(\|\bu\|)}{h})f(\br(\bu))\sqrt{\det(g(\bu))}{\rm d}\bu
	\end{aligned}
	\end{equation}
	After the changing of variable $\bu=h\bz$, note that by symmetry we have
	\begin{equation}
	\int_{\|\bz\|\le 1}z^l\|\bz\|^2K(\|\bz\|)d\bz=0
	\end{equation}
	Thus the coefficient of $h^3$ vanishes. Set 
	\begin{equation}
	\begin{aligned}
	&\partial_l\eta^k(0)=\left.\frac{\eta^k(\br(\bu))}{\partial u^l}\right|_{\bu=0}\\
	&\partial_j\partial_sr^i(0)=\left.\frac{\partial r^i(\bu)}{\partial u^j\partial u^s}\right|_{\bu=0}
	\end{aligned}
	\end{equation}
	The leading term is 
	\begin{equation}
	h^4\left(\bigg(\partial_l\eta^k(0)+\frac{1}{2}\eta^k(x)\sum_{i,j}\partial_j\partial_jr^i(0)\partial_lr^i(0)\bigg)f(x)\int_{\|\bz\|\le 1}(z^1)^2\|\bz\|^2K(\|\bz\|){\rm d}\bz\right)
	\end{equation}
	Utilizing the estimation for $\bd$, we obtain that 
	\begin{equation}
	\|\what{A}-A\|_F^2\le \|\ef[\bt\bx^t\bk_P]\|_F^2\|\bd^{-1}\|_F^2=O(h^4)
	\end{equation}
\end{proof}

For the matrix $\wtid{D}$ we have error coming from random sampling. Therefore, the estimation only holds with high probability (w.h.p.). That is, for any $\epsilon>0$ the estimation holds with probability greater than $1-\epsilon$. For the simplicity of statements, we omit the specific computation of quantities involving $\epsilon$. 
\begin{lemma}\label{estimatetideD}
	Suppose that when $h\to 0$ and $nh^m\to \infty$ as $n\to \infty$, w.h.p. the following equality holds 
	\begin{equation}\label{wtidequalsbd}
	\wtid{D}=\bd+O(\frac{h^2}{\sqrt{nh^m}})
	\end{equation}
	Thus the inverse is given by
	\begin{equation}
	\wtid{D}^{-1}=\bd^{-1}-\bd^{-1}O(\frac{h^2}{\sqrt{nh^m}})\bd^{-1}
	\end{equation}
\end{lemma}
\begin{proof}
	For an arbitrary element $\wtid{D}_{kl}$, note that
	\begin{equation}
	\ef[\wtid{D}_{kl}]=\bd_{kl}
	\end{equation} 
	The variance is 
	\begin{equation}
	\ef\left[(\wtid{D}_{kl}-\bd_{kl})^2\right]=\frac{1}{n}\ef\left[((\bx)_k(\bx)_l\bk_X-\bd_{kl})^2\right]\le \frac{1}{n}\ef\left[((\bx)_k(\bx)_l\bk_P)^2\right]
	\end{equation}
	Using the same method as in the proof in lemma \ref{estimatebd-lemma}, we can find the leading term is 
	\begin{equation}
	\frac{1}{h^m}\left(h^4f(x)\int_{\|\bz\|\le 1}(z^1)^2(z^2)^2K^2(\|\bz\|){\rm d}\bz+o(h^4)\right)
	\end{equation}
	By Chebyshev's inequality, for any $\epsilon>0$,
	\begin{equation}
	\mathbb{P}\left(|\wtid{D}_{kl}-\bd_{kl}|\ge\epsilon\right)\le\frac{1}{\epsilon^2}O(\frac{h^4}{nh^m})
	\end{equation}
	as $h\to 0$ and $nh^m\to\infty$. Set the right side to be $O(1)$, we can show that w.h.p. $|\wtid{D}_{kl}-\bd_{kl}|\le h^2/\sqrt{nh^m}$, which proves the equation \eqref{wtidequalsbd}. The inverse is given by the following identity 
	\begin{equation}
	\begin{aligned}
	(\bd+O(\frac{h^2}{\sqrt{nh^m}}))^{-1}&=(I+\bd^{-1}O(\frac{h^2}{\sqrt{nh^m}}))^{-1}\bd^{-1}\\
	&=(I-\bd^{-1}O(\frac{h^2}{\sqrt{nh^m}}))\bd^{-1}\\
	&=\bd^{-1}-\bd^{-1}O(\frac{h^2}{\sqrt{nh^m}})\bd^{-1}
	\end{aligned}
	\end{equation}
\end{proof}
Using the presented results, now we can estimate the variance. 
\begin{lemma}\label{variance-lemma}
	Suppose that when $h\to 0$ and $nh^m\to \infty$ as $n\to \infty$, the variance is $O(\frac{1}{nh^m})$ w.h.p.
\end{lemma}
\begin{proof}
	First by interpolating a mixing term we have
	\begin{equation}
	\begin{aligned}
	&\ef\left[\|\mathbf{A}-\wtid{A}\|_F^2\right]=	\ef\left[\|\mathbf{A}-\wtid{L}\bd^{-1}+\wtid{L}\bd^{-1}-\wtid{A}\|_F^2\right]\\
	\le& 2\bigg(\underbrace{\ef\left[\|\mathbf{A}-\wtid{L}\bd^{-1}\|_F^2\right]}_{(*)}+\underbrace{\ef\left[\|\wtid{L}\bd^{-1}-\wtid{A}\|_F^2\right]}_{(**)}\bigg)
	\end{aligned}
	\end{equation}
	For the first term we have
	\begin{equation}
	(*)=\ef\left[\|(\bl-\wtid{L})\bd^{-1}\|_F^2\right]\le\ef\left[\|\bl-\wtid{L}\|_F^2\right]\|\bd^{-1}\|_F^2
	\end{equation}
	For an arbitrary term in $\bl-\wtid{L}$ we have
	\begin{equation}\label{variance1}
	\begin{aligned}
	&\ef\left[\bigg(\frac{\sum_{i=1}^{n}(\Xi_i)_k(\Delta_i)_lK_{x_i}}{n}-\ef[(\bxi)_k(\bx)_l\bk_P]\bigg)^2\right]\\
	=&\frac{1}{n}\mathbb{V}\text{ar}_f\left[(\bxi)_k(\bx)_l\bk_P\right]\le\frac{1}{n}\ef\left[(\bxi)_k^2(\bx)_l^2\bk^2_P\right]
	\end{aligned}
	\end{equation}
	When $h<h_0$, by the model assumption \eqref{modelequation}, the above quantity \eqref{variance1} is bounded by
	\begin{equation}
	\begin{aligned}
	\frac{\|A\|^2_F}{h^{2m}}\int_{\|\bu\|\le R_x(h)}(u^k+o(\|\bu\|))^2(u^l+o(\|\bu\|))^2K^2(\frac{\|\bu\|+o(\|\bu\|)}{h})f(\br(\bu))\sqrt{\det(g(\bu))}{\rm d}\bu
	\end{aligned}
	\end{equation}
	where the leading term is 
	\begin{equation}
	f(x)\frac{\|A\|_F^2}{h^{m-4}}\int_{\|\bz\|\le 1}(z^1)^2(z^2)^2K(\|\bz\|)^2{\rm d}\bz
	\end{equation}
	Together with the estimation for $\bd^{-1}$, we see that $(*)= O(\frac{1}{nh^m})$.
	
	For the second term we have
	\begin{equation}\label{variance2}
	\begin{aligned}
	(**)&=\ef\left[\|\wtid{L}(\bd^{-1}-\wtid{D}^{-1})\|_F^2\right]=\ef\left[\|\wtid{L}\bd^{-1}(\wtid{D}-\bd)\wtid{D}^{-1}\|_F^2\right]\\
	&\le \ef\left[\|\wtid{L}\|_F^2\right]\ef\left[\|\wtid{D}-\bd\|_F^2\right]\ef\left[\|\wtid{D}^{-1}\|_F^2\right]\|\bd^{-1}\|_F^2
	\end{aligned}
	\end{equation} 
	Now the order of each quantity in \eqref{variance2} is 
	\begin{itemize}
		\item[(1)] Similar to the proof of lemma \ref{estimatetideD}, note that $\ef[\wtid{L}_{kl}]=\bl_{kl}$ and the variance is controlled by a quantity of order $O(h^4/nh^m)$ w.h.p. Thus,
		\begin{equation}
		\ef\left[\|\wtid{L}\|_F^2\right]\le O(h^4)
		\end{equation}
		\item[(2)] Using the same method as in the estimation for $\|\bl-\wtid{L}\|_F^2$, we see that it is controlled by $O(1/nh^{m-4})$ w.h.p.
		\item[(3)] By lemma \ref{estimatetideD}, this term is $O(1/h^4)$ w.h.p.
		\item[(4)] By lemma \ref{estimatebd-lemma}, this term is $O(1/h^4)$.
	\end{itemize}
	Hence, the order of $(**)$ is $O(\frac{1}{nh^m})$. Overall, the rate for variance is proved.
\end{proof}

Altogether we can give the convergence rate of MSE.
\begin{theorem}\label{mse-rate-framed-theorem}
	Let $\xi$ be a normal vector field on $\mm$, and $P$ be a random vector valued in $\mm$ with smooth positive density function $f$. Assume that $K:\mathbb{R}\to \mathbb{R}$ is a twice differentiable function supported on $[0,1]$.  In addition,  assume that a basis $e_1,e_2,\cdots,e_m$ for the tangent space $\txm$ is given. Let $x_1,\cdots, x_n$ be i.i.d. samples from $P$.  When $h\to 0$ and $nh^m\to \infty$ as $n\to \infty$, the mean square error defined as in \eqref{msedefinition} is 
	\begin{equation}
	\mathrm{MSE}= O(h^4)+O(\frac{1}{nh^m})
	\end{equation} 
	If $h$ is chosen to be proportional to $n^{-1/(m+4)}$, the optimal convergence rate is given by $O(n^{-4/(m+4)})$.
\end{theorem}

\subsection{$k$-Nearest-Neighbor Method}
$k$-nearest-neighbor method is widely used in many settings since it is intuitively simple, easy to implement and computationally efficient. For simplicity, we focus on disk neighbors with fixed radius $h$ in this article. However, the disk neighbor can be replaced by $k$-nearest-neighbor without any other changes. According to Theorem \ref{mse-rate-framed-theorem} and the well known relation $\frac{k}{n} \sim h^m$ (citation), we suggest to set $k = O(n^{4/(m+4)})$ in practice to reach the optimal convergence rate $O(n^{-4/(m+4)})$.  

\subsection{Comparison with Other Estimators}

In \cite{buet2018discretization}, the authors proposed generalized second fundamental form and mean curvature for varifolds based on geometric measure theory. Therefore, even if the underlying space has singularities (for example, self-intersecting points), they are able to compute the generalized mean curvature. Basically, a $m$-varifold is a Radon measure on the space $\mathbb{R}^d\times \mathbb{G}_{m,d}$ where $\mathbb{G}_{m,d}$ is the Grassmannian manifold of $m$-planes in $\mathbb{R}^d$. For an $m$-varifold $V$, they define the generalized mean curvature field using the first variation
\begin{equation}
	\begin{aligned}
	\delta V: C_c^1(\mathbb{R}^d,\mathbb{R}^d)&\to \mathbb{R}\\
	 X&\to \int_{\mathbb{R}^d\cap\mathbb{G}_{m,d}}\text{div}X(x){\rm d}V(x,S)
	\end{aligned}
\end{equation}
The generalized mean curvature descends to the classical mean curvature in the sense that $\delta V=-H\mathcal{H}^m_{|\mm}$ if $V$ is the standard measure on $\mm$.  However, in general, the first variation cannot be given in closed form. For a point cloud varifold $V=\sum_j=1^n m_j\delta_{x_j}\otimes \delta_{P_j}$, the first variation $\delta V$ is not a measure. Therefore, they put forward the following quantity to approximate the mean curvature for point clouds
\begin{equation}
	H^V_{\alpha,\beta,\epsilon}=\frac{C_{\beta}}{C_{\alpha}\epsilon}\frac{\sum_{x_j\in B_\epsilon(x)\backslash\{x\}}m_j\alpha'(\frac{\|x_j-x\|}{\epsilon})\Pi_{P_j}\frac{x_j-x}{\|x_j-x\|}}{\sum_{x_j\in B_{\epsilon}(x)}m_j\beta(\frac{\|x_j-x\|}{\epsilon})}
\end{equation} 
where $\alpha,\beta$ are functions supported on $[-1,1]$ and such that $\int_{\mathbb{R}^d}\alpha(x){\rm d}x=\int_{\mathbb{R}^d}\beta(x){\rm d}x=1$, and
$C_\alpha,C_\beta$ are constants related to $\alpha,\beta$. They proved that $H^V_{\alpha,\beta,\epsilon}$ converges to $H$ under certain conditions (Theorem 3.6 in \cite{buet2018discretization}). Specifically, if the points are uniformly sampled and the tangents are exact, the convergence rate is $\frac{1}{n\epsilon}+\epsilon$ given that $n\epsilon\to\infty$. Therefore, the optimal bandwidth is $\epsilon=n^{-1/2}$ and the optimal convergence rate is also $n^{-1/2}$. Compared with WME algorithm, this method yields faster convergence when the underlying manifold is of relatively high dimension. However, for curves and surfaces, our algorithm provides faster convergence. 

In \cite{aamari2019nonasymptotic}, the authors proposed estimators for the tangent space, the second fundamental form and the support of manifold based on local polynomial fitting. Finite sample rates are derived. Let $\mm$ be a submanifold subject to some regularity conditions, and in addition assume that the data are sampled uniformly on the manifold. Then the convergence rate is $O(n^{-(k-2)/m})$ where $k$ is regularity parameter and $m$ is the dimension of manifold. However, the estimator cannot be given in a closed form. Nonconvex optimization techniques should be involved in order to approximate the estimators.

\section{Numerical Experiments}\label{numericalexperiments}

We apply WME algorithm on various synthetic data sets and verify the optimal convergence rate proved in theorem \ref{mse-rate-framed-theorem}. Both kernel method and $k$-Nearest-Neighbor method are used to show the consistency of WME algorithm. Then the algorithm is applied to curvature estimation. A comparison with classical local quadratic fitting method is carried out to demonstrate the efficiency and robustness.

\subsection{Kernel Method}
We verify the convergence rates on three synthetic data sets: 

\textbf{Conical Spiral: } A conical spiral is a space curve given by 
\begin{equation}
\gamma(t) = (rt\cos(at),rt\sin(at),t)
\end{equation}
where $r,a>0$ are parameters. Here we set $r=1$ and $a=1.5$. It is a 1-dimensional manifold embedded in 3-Euclidean space. If we choose $h$ proportional to $n^{-1/5}$, then the optimal convergence rate will be $O(n^{-4/5})$.

\textbf{Torus: } A 2-dimensional torus is given by
\begin{equation}
F(\theta,\alpha)=(R+r\cos(\theta)\cos(\alpha),(R+r\cos(\theta))\sin(\alpha),r\sin(\theta))
\end{equation} 
where $R>2r>0$ are parameters. Here we set $R=4$ and $r=0.5$. Thus if we choose $h$ proportional to $n^{-1/6}$, the optimal rate will be $O(n^{-2/3})$.

\textbf{Ellipsoid: } A 3-dimensional ellipsoid is given by
\begin{equation}
F(\theta,\alpha,\beta)=(a\cos(\theta),b\sin(\theta)\cos(\alpha),c\sin(\theta)\sin(\alpha)\cos(\beta),d\sin(\theta)\sin(\alpha)\sin(\beta))
\end{equation}
where $a,b,c,d>0$ are parameters. Here we set $a=1$, $b=1.15$, $c=1.31$ and $d=1.7$. If the bandwidth is chosen proportional to $n^{-1/7}$, the convergence rate will be $O(n^{-4/7})$.

\begin{figure}[htbp]
	\centering
	\begin{subfigure}{0.3\textwidth}
		\includegraphics[width=\textwidth]{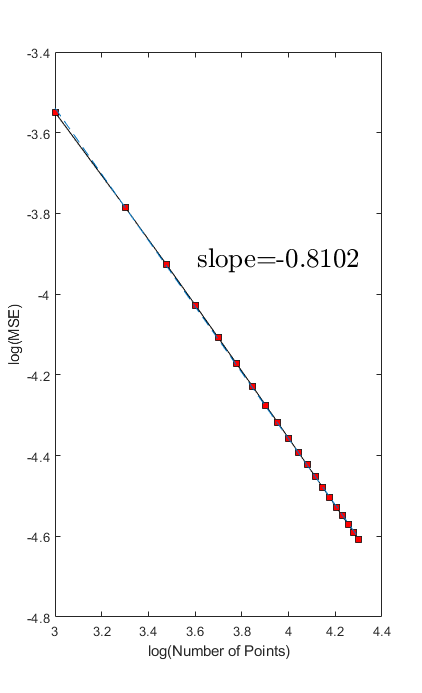}
		\caption{Conical Spiral}
	\end{subfigure}
	\begin{subfigure}{0.3\textwidth}
		\includegraphics[width=\textwidth]{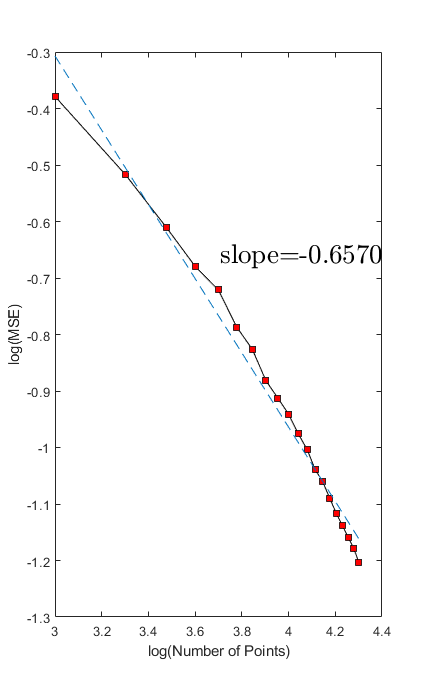}
		\caption{Torus}
	\end{subfigure}
	\begin{subfigure}{0.3\textwidth}
		\includegraphics[width=\textwidth]{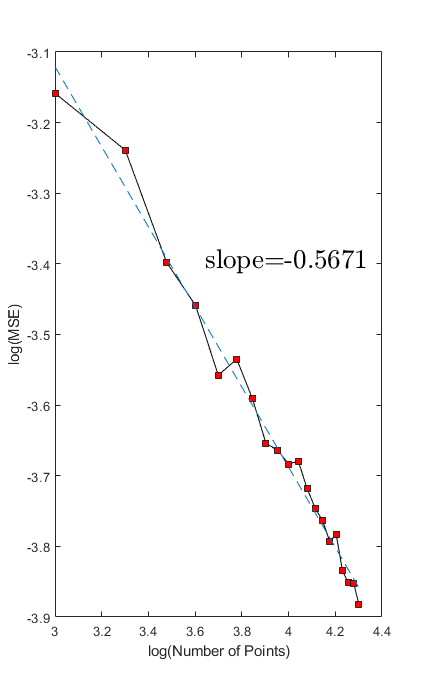}
		\caption{Ellipsoid}
	\end{subfigure}
	\caption{The $\log(n)$-$\log(\mse)$ plot for conical spiral, 2-dimensional torus and 3-dimensional ellipsoid.  Slopes match the rates of convergence.}
	\label{convergence rate}	
\end{figure}

We uniformly sample $1,000\sim 20,000$ points using the parametrizations given as above. The bandwidth is chosen so that the convergence rate will be optimal. We perform leave-one-out estimates on the data points. The $\log(\text{MSE})$ is drawn with respect to $\log(\text{number of points})$. Therefore the slope corresponds to the order of convergence rate. The results are shown in figure \ref{convergence rate}.

\subsection{$k$-Nearest-Neighbor Method}

\begin{figure}[htbp]
	\centering
	\begin{subfigure}[b]{0.3\textwidth}
		\includegraphics[width=\textwidth]{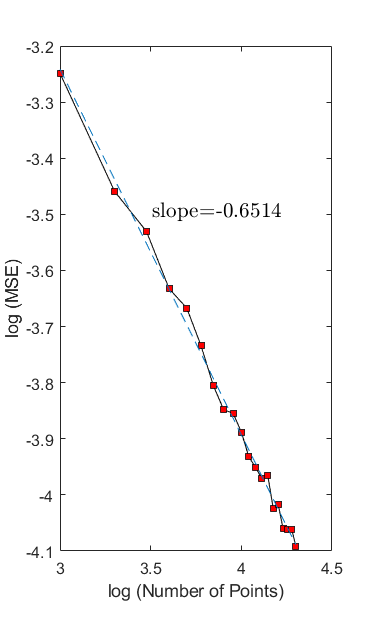}
		\caption{Torus}
	\end{subfigure}
	\begin{subfigure}[b]{0.3\textwidth}
		\includegraphics[width=\textwidth]{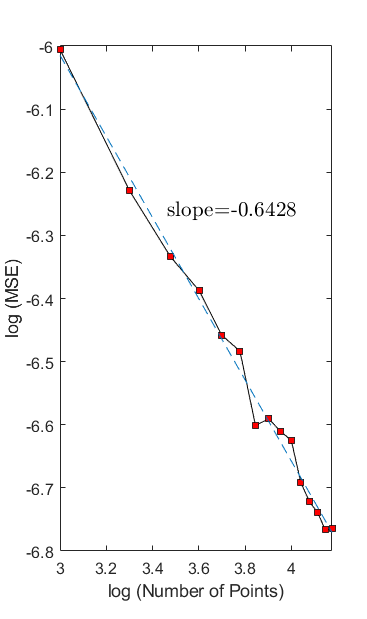}
		\caption{Ellipsoid}
	\end{subfigure}
	\caption{The $\log(n)$-$\log(\mse)$ plot for 2-dimensional torus and 2-dimensional ellipsoid. Slopes match the convergence rate.}\label{rate}
\end{figure}

We test the optimal convergence rate on a 2-dimensional torus and a 2-dimensional ellipsoid. The sample size for the torus with major radius $5$ and minor radius $2$ ranges from $1000$ to $20000$. The sample size for the ellipsoid with length of principal axes $6,6,8$ ranges from $7000$ to $30000$. $k$ is chosen to be $n^{\frac{2}{3}}$. Points are uniformly sampled. Figure \ref{rate} is the $
\log (n)$-$\log(\mse)$ plot. The slopes match the optimal convergence rate.

\subsection{Curvature Estimation}

    \begin{figure}[htbp]
	\centering
	\begin{subfigure}[b]{0.4\textwidth}
		\includegraphics[width=\textwidth]{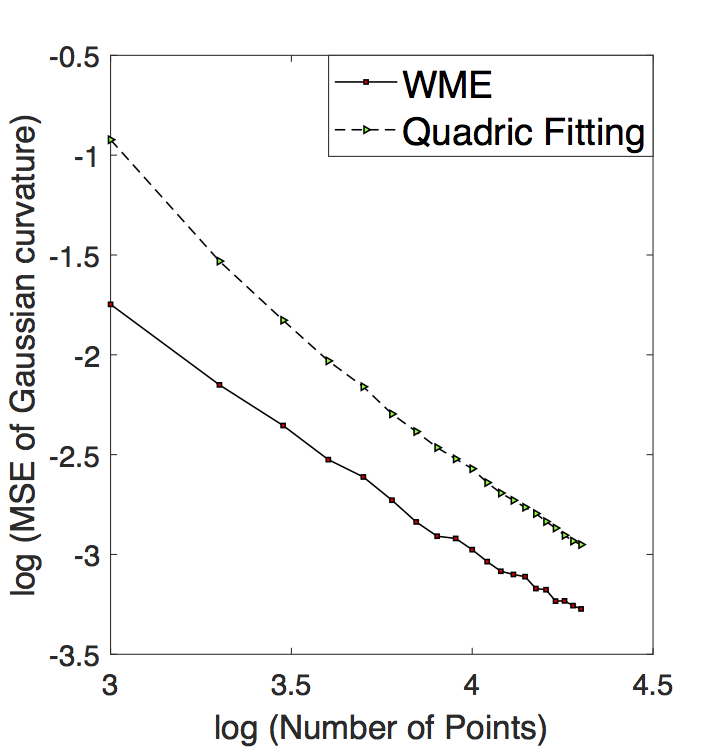}
		\caption{Gaussian Curvature}
	\end{subfigure}
	\begin{subfigure}[b]{0.4\textwidth}
		\includegraphics[width=\textwidth]{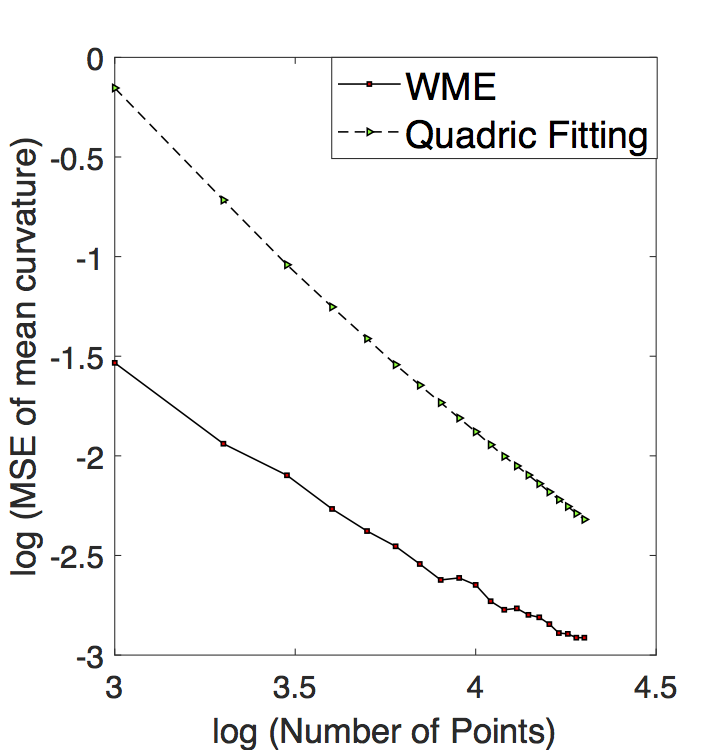}
		\caption{Mean Curvature}
	\end{subfigure}
	\caption{MSE of Gaussian curvature and mean curvature on the torus obtained by WME and quadratic fitting.}\label{torus}
\end{figure}

We compare our method with a local quadratic surface fitting method. This is chosen for two reasons. On one hand, quadratic fitting is a commonly used method. Other complicated fitting algorithms involve extra scaling parameters which are difficult to tune in practice. On the other hand, quadratic fitting is studied by many scientists.
In \cite{AcomparisonofGaussianandmeancurvatureestimation}, the authors compared five  methods in computing Gaussian and mean curvature for triangular meshes of 2 dimensional surfaces.  The result turns out that quadratic fitting exceeds other methods in computing mean curvature.

  The method of quadratic surface fitting is illustrated as follows. First translate and rotate the $k$-nearest neighbors of a point so that its normal vector coincides with $z$-axis.
  Then fit the paraboloid $z=ax^2+bxy+cy^2$ by least square. The Gaussian curvature and mean curvature at origin $P$ are given by
  \begin{equation}
    K=4ac-b^2,H=a+c.
  \end{equation}
  The MSE of Gaussian curvature and mean curvature are compared as follows. We sample $1000\sim 20000$ points on a torus with major radius 5 and minor radius 2. The number of $k$-nearest neighbors is set to be 100 for each iteration. The result in Figure 2 shows that our method excels the quadratic fitting method without introducing any computational complexity.

  \begin{figure}[htbp]
  \centering
  \begin{subfigure}[b]{0.4\textwidth}
  \includegraphics[width=\textwidth]{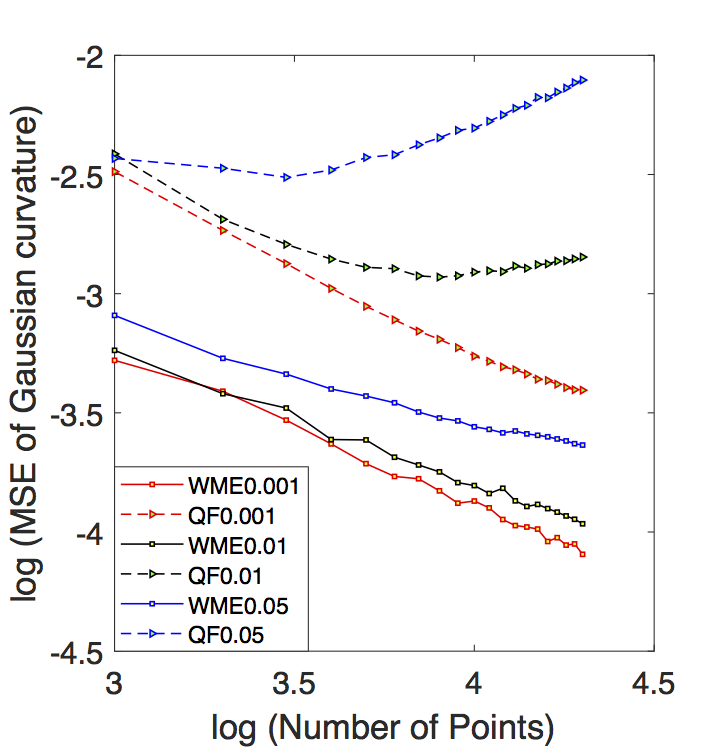}
  \caption{Gaussian Curvature}
  \end{subfigure}
  \begin{subfigure}[b]{0.4\textwidth}
  \includegraphics[width=\textwidth]{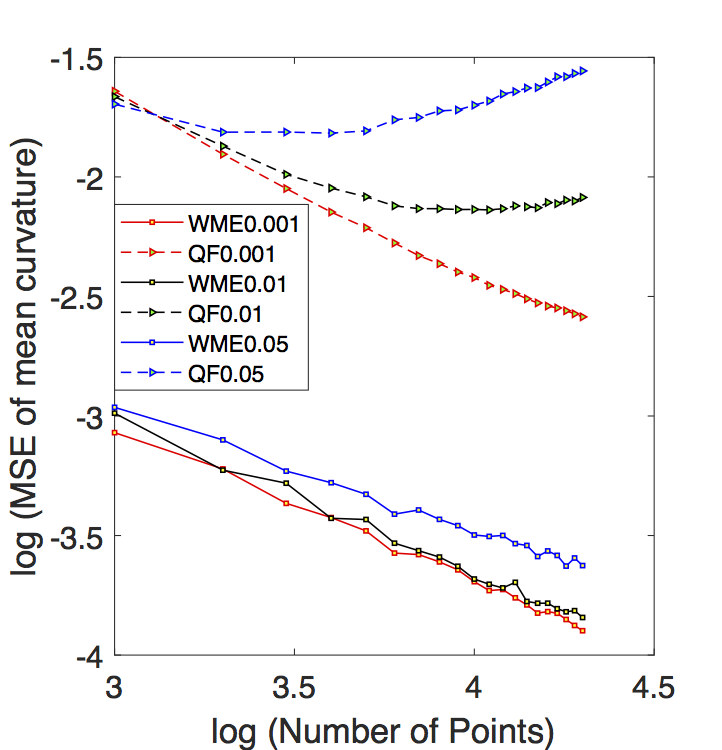}
  \caption{Mean Curvature}
  \end{subfigure}
  \caption{Robustness comparison on the noisy torus. Left and right plots are comparisons of Gaussian and mean curvature respectively.}\label{noisytorus}
  \end{figure}

  The robustness is compared as follows. Again we sample $1000,2000,\cdots,20000$ points on the same torus with multivariate Gaussian noise with zero mean and covariance $\sigma^2 I_3$ where $\sigma^2\in\{0.01,0.05,0.001\}$. The MSE of Gaussian curvature and mean curvature for different noises in Figure \ref{noisytorus} show that our method is more robust.

\section{Applications}\label{simplification}

We apply our WME algorithm to real data sets. The first application is curvature estimation for brain cortical surfaces. We demonstrate the robustness of WME method on a real cortical surface data set. The second application is point cloud simplification which is a hot topic in computer vision. We propose a new method based on curvature to simplify large point cloud data sets.  The results of following experiments show that our algorithm is also practical for real point clouds.

\begin{figure}
  \centering
     \begin{subfigure}[b]{0.3\textwidth}
  \includegraphics[width=\textwidth]{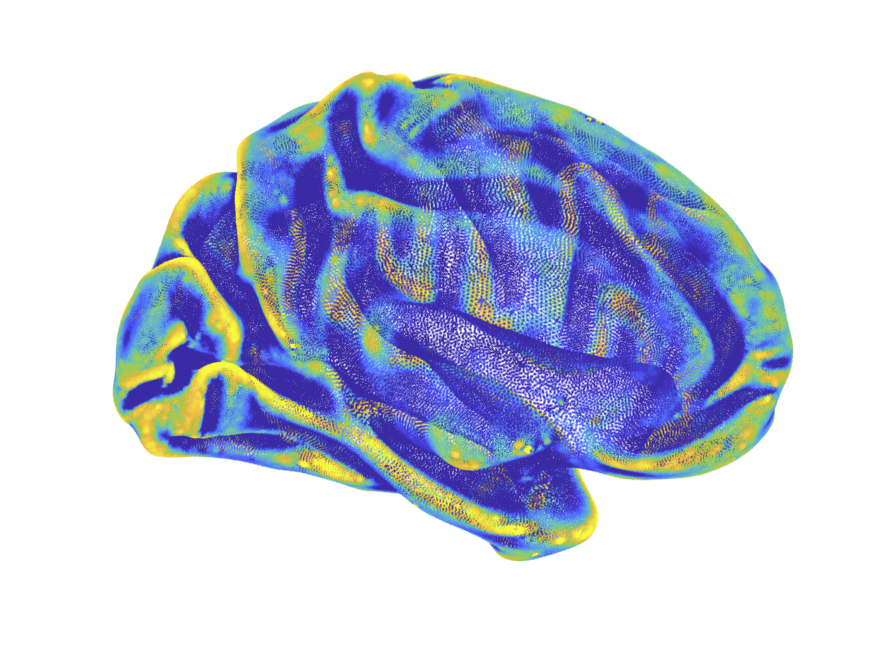}
  \caption{Cortical Surface}
  \end{subfigure}
    \begin{subfigure}[b]{0.3\textwidth}
  \includegraphics[width=\textwidth]{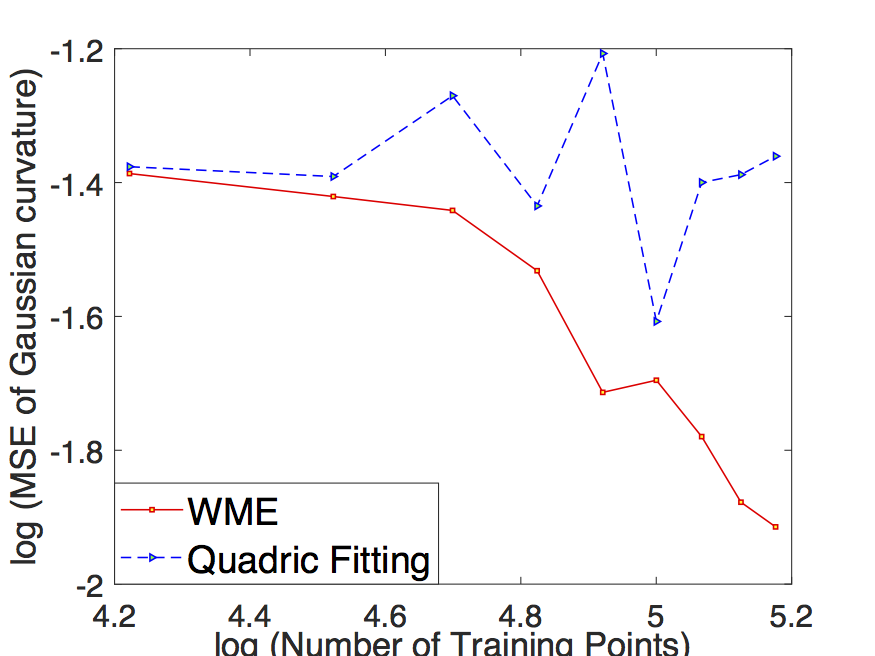}
  \caption{Gaussian Curvature}
  \end{subfigure}
    \begin{subfigure}[b]{0.3\textwidth}
  \includegraphics[width=\textwidth]{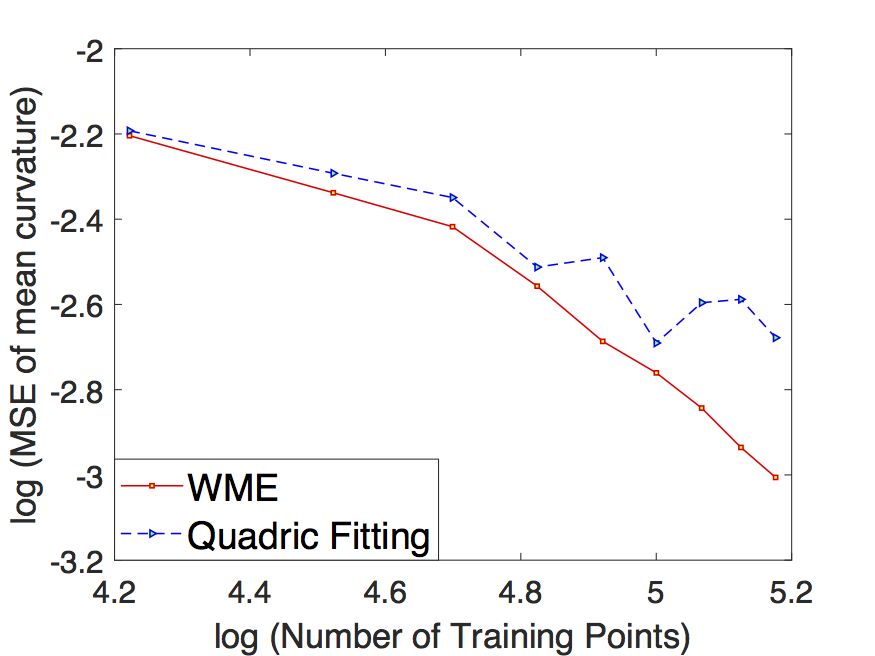}
  \caption{Mean Curvature}
  \end{subfigure}
  \caption{The first panel is the cortical surface data colored by the Gaussian curvature. The last two plots are $\log(\mse)$ of the Gaussian and the mean curvature on this dataset.}\label{crossvalidation}
\end{figure}

\subsection{Brain Cortical Surface Data}
To further illustrate the robustness, we test our method on the real brain cortical surface data. A point cloud of left brain cortical surface is obtained from Human Connectome Projects (\url{s3://hcp-openaccess/HCP_1200/100206/MNINonLinear/Native/}), consisting of 166,737 position vectors. This data is noisy and there is no information about the true curvature of the surface so there is no ground truth and the error can't be calculated. Instead we propose an indirect way to evaluate the performance. Firstly, we estimate the Gaussian and mean curvature for the point cloud based on the entire dataset. The results are regarded as the true curvature for the underlying cortical surface. Then, the data is divided into training and testing sets. We recalculate the Gaussian and mean curvature for training data. For each testing data, the curvature is inferred to be the mean of curvature for its $k$ nearest neighbors in training data. Finally, we compute the mean square error of the curvature for testing data. The same procedure is also carried out using quadratic surface fitting method. From Figure \ref{crossvalidation}, the mean square error obtained from WME is monotonically decreasing as the number of testing data increases but the error from quadratic surface fitting method is fluctuating, which means that WME is more robust on this real and complicated dataset.


%
%
%
%

\subsection{Point Cloud Simplification}

Point clouds are often converted into a continuous surface representation such as polygonal meshes and splines. This process is called surface reconstruction \cite{Berger2016A}. The reconstruction algorithms require large amounts of memory and do not scale well with data size. Hence before further processed, the complexity of point cloud data should be reduced first. In \cite{Efficientsimplification}, the authors proposed three types of simplification algorithms: clustering methods, iterative simplification and particle simulation. These methods are based on a quantity defined by the covariance of local data. As claimed by \cite{Efficientsimplification}, this quantity reflects the curving of point cloud. However, the clear relation between this quantity and the curvatures needs to be further studied. Here we propose a curvature-adaptive clustering simplification algorithm and compare with uniform clustering simplification algorithm.

The uniform clustering method is described as follows. Starting from a random seed point, a cluster $C_0$ is built by successively adding the nearest neighbors. The process is terminated when the size of clusters reaches the previously set threshold. The next cluster $C_1$ is built in the same procedure with all points in $C_0$ excluded. Each cluster is represented by its mean as a representative. The simplified point cloud is given by the representatives.

Intuitively, to preserve the geometric details of point cloud, the points in highly curved region should be kept. Therefore, a seed point with larger curvature should generate smaller cluster. Let $\Omega$ represent any kind of (Gaussian, mean or principal) curvature. Suppose that $|\Omega|_{\max}$ is the largest absolute curvature of the entire surface. Starting from a random seed point $p$, with absolute curvature $|\Omega|_p$, a cluster $C_p$ is built by successively adding the nearest neighbors. The process is terminated when the size of cluster reaches
  \begin{equation}
    \#C_p=\lceil(1-c\frac{|\Omega_p|}{|\Omega|_{\max}})T\rceil
  \end{equation}
where $0<c<1$ is the scaling constant and $T$ is the preset threshold. $\lceil\cdot\rceil$ denotes the ceiling function. The cluster and its curvature are represented by the mean of its points and mean of corresponding curvature. This yields a non-uniform clustering method.

\begin{figure}[h]

\centering
\begin{subfigure}[b]{0.2\textwidth}
  \includegraphics[width=\textwidth]{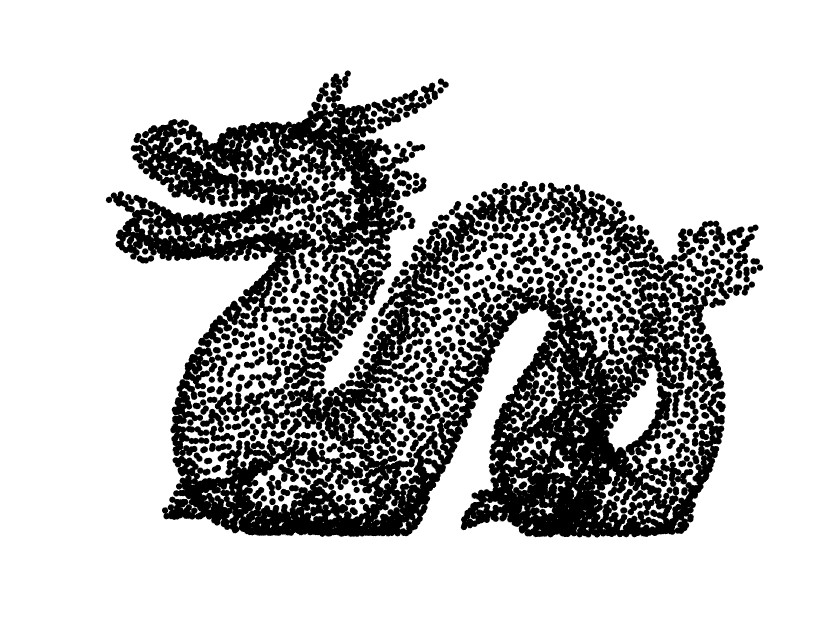}
\end{subfigure}
\begin{subfigure}[b]{0.2\textwidth}
  \includegraphics[width=\textwidth]{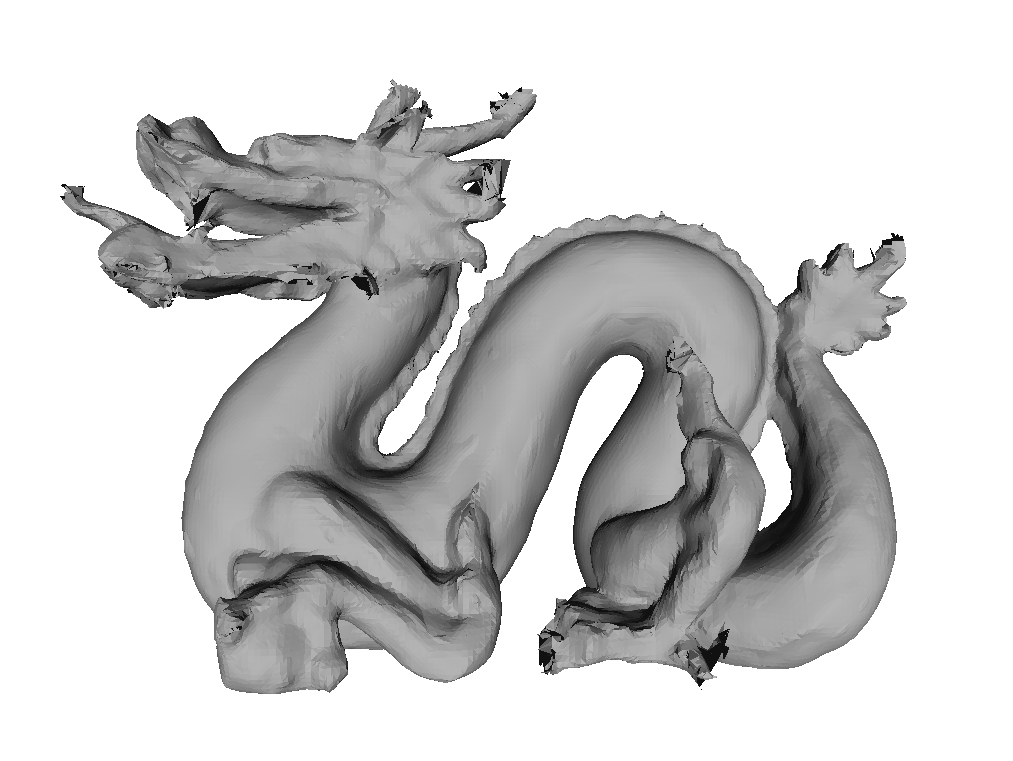}
\end{subfigure}
\begin{subfigure}[b]{0.2\textwidth}
  \includegraphics[width=\textwidth]{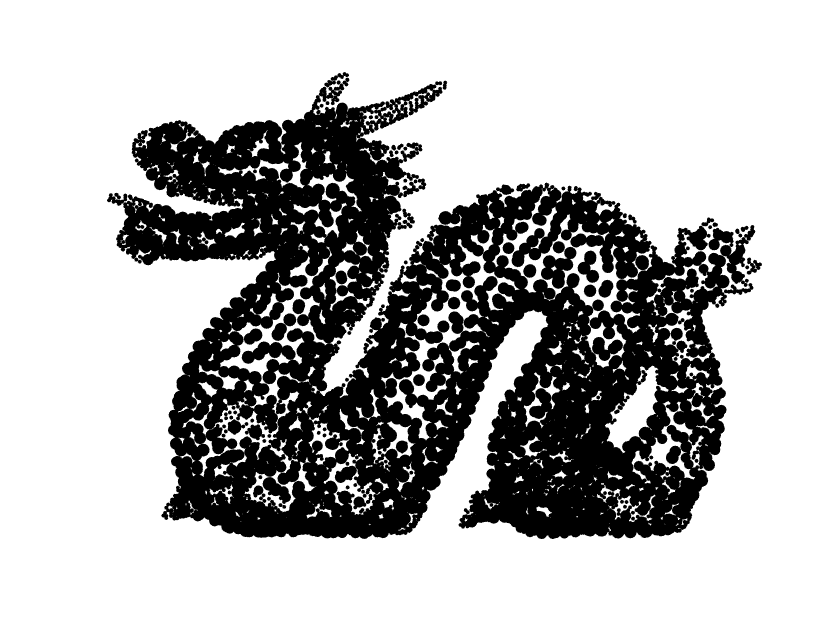}
\end{subfigure}
\begin{subfigure}[b]{0.2\textwidth}
  \includegraphics[width=\textwidth]{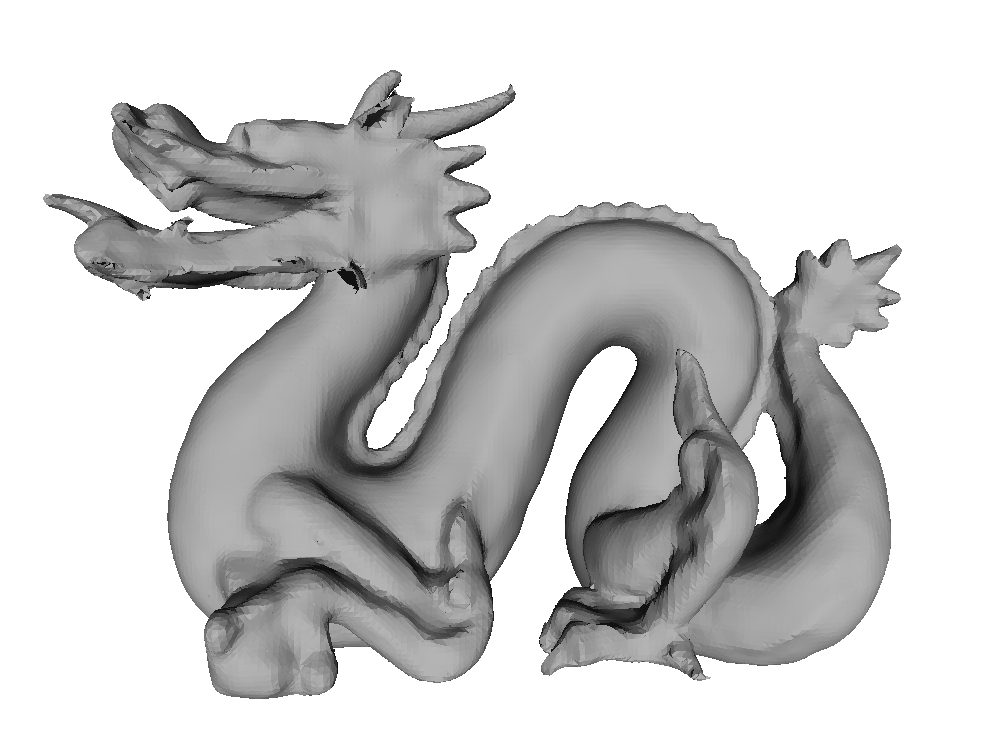}
\end{subfigure}


  \caption{Duke Dragon dataset. The surfaces are reconstructed from 6500 points.}\label{Dragon}
  \end{figure}
\begin{figure}[h]
\centering

\begin{subfigure}[b]{0.22\textwidth}
  \includegraphics[width=\textwidth]{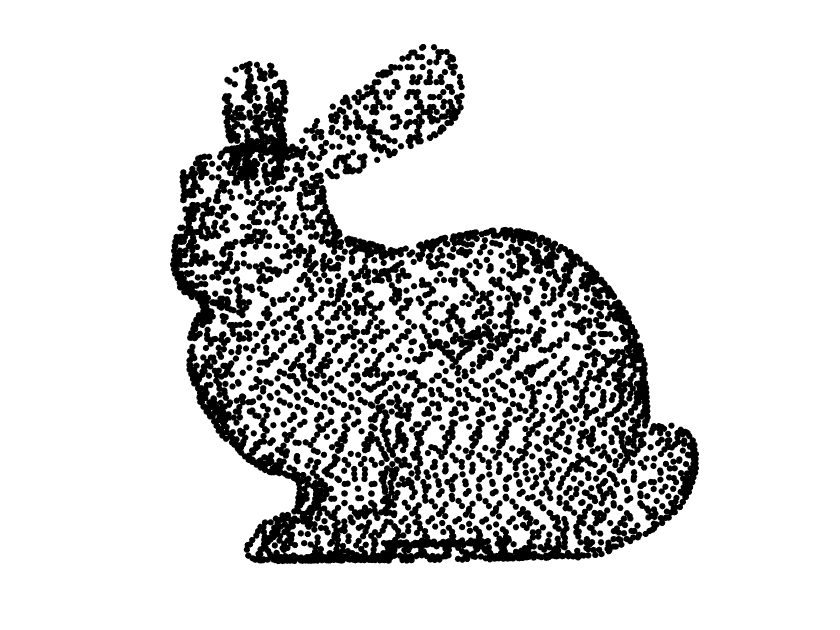}
\end{subfigure}
\begin{subfigure}[b]{0.18\textwidth}
  \includegraphics[width=\textwidth]{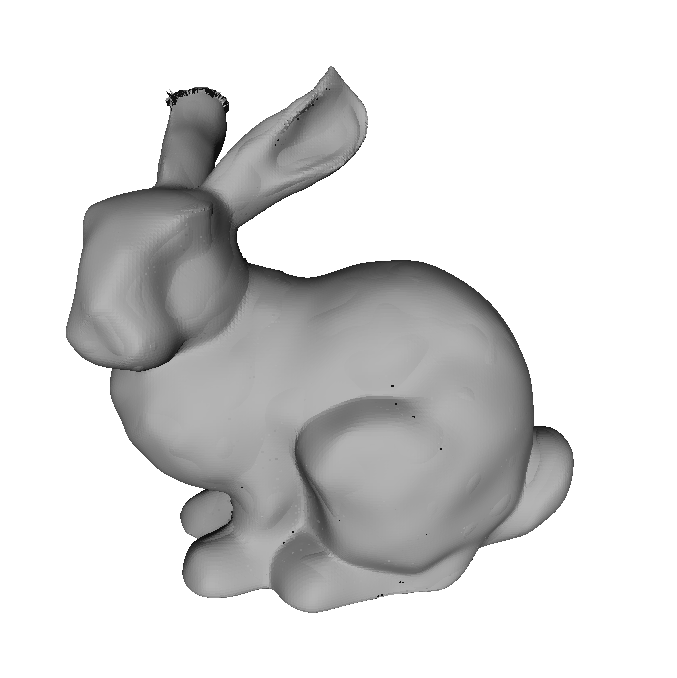}
\end{subfigure}
\begin{subfigure}[b]{0.22\textwidth}
  \includegraphics[width=\textwidth]{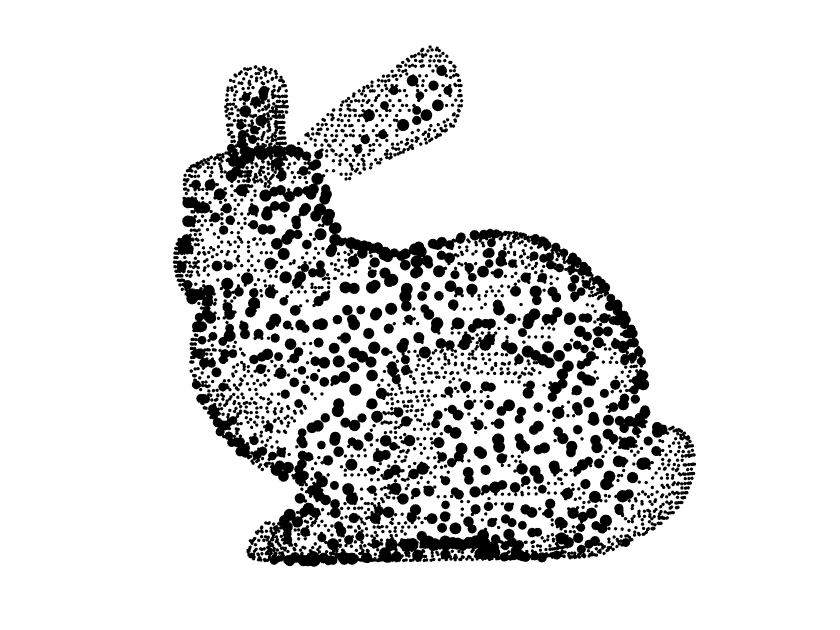}
\end{subfigure}
\begin{subfigure}[b]{0.18\textwidth}
  \includegraphics[width=\textwidth]{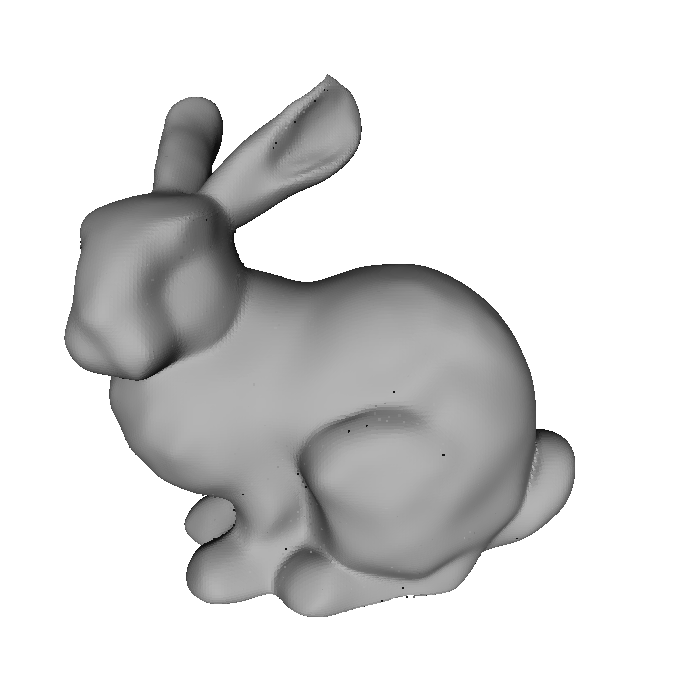}
\end{subfigure}


  \caption{Stanford Bunny dataset. The surfaces are reconstructed from 4400 points.}\label{Bunny}
  \end{figure}
\begin{figure}[h]
  \centering
  \begin{subfigure}[t]{0.24\textwidth}
  \includegraphics[width=\textwidth]{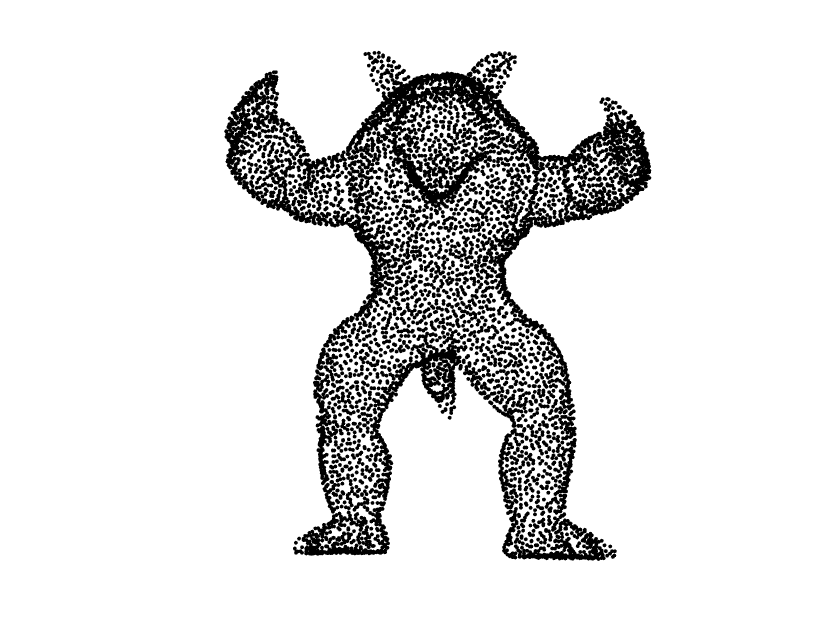}
\end{subfigure}
\begin{subfigure}[t]{0.16\textwidth}
  \includegraphics[width=\textwidth]{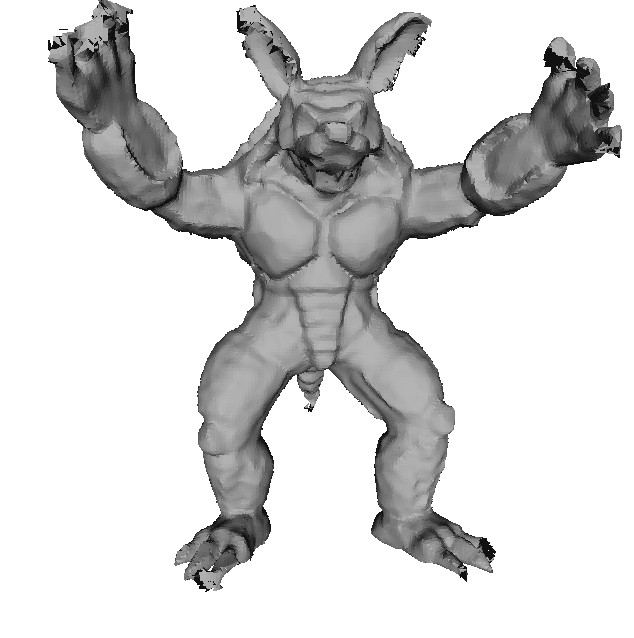}
\end{subfigure}
\begin{subfigure}[t]{0.24\textwidth}
  \includegraphics[width=\textwidth]{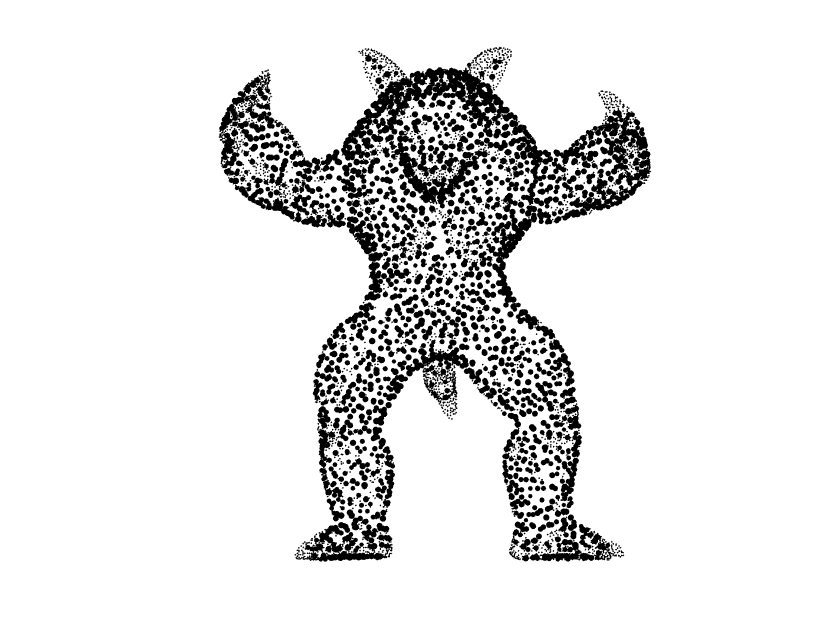}
\end{subfigure}
\begin{subfigure}[t]{0.16\textwidth}
  \includegraphics[width=\textwidth]{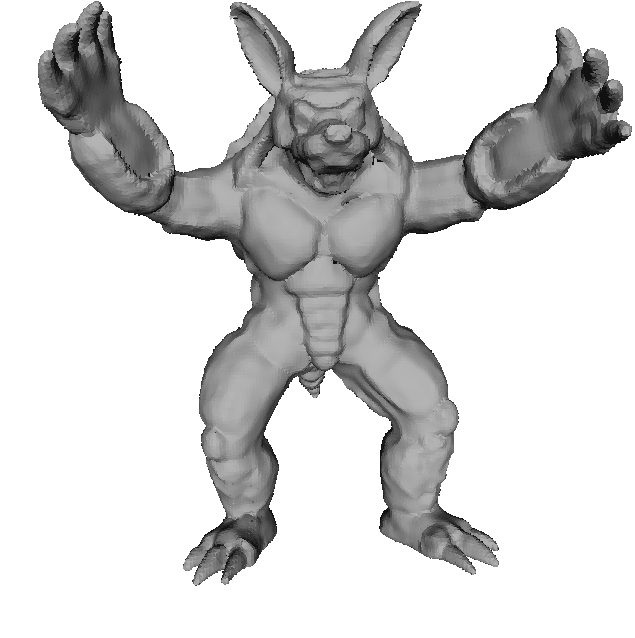}
\end{subfigure}


  \caption{Armadillo dataset. The surfaces are reconstructed from 7800 points.}\label{Armadillo}
\end{figure}

The algorithms are applied to three scanned data sets: the Duke dragon, the Stanford bunny and the Armadillo. Here we adopt absolute mean curvature for curvature-based simplification.
 After simplification, we apply the Moving Least Square (MLS) method for surface reconstruction \cite{Berger2016A}. The visualized surfaces in Figure \ref{Dragon}, \ref{Bunny} and \ref{Armadillo} give a direct comparison of two algorithms. In each of these figures, the first subfigure represents the simplified point cloud using uniform clustering and the second subfigure is the reconstructed surface from the point cloud. Similarly, the third subfigure represents result using curvature-based clustering and the fourth subfigure shows the surface reconstructed from it.  Results show that WME preserves more geometric information than the uniform method, especially for the region with larger curvature.


\section{Discussions and Future Works}\label{discussion}

This paper introduced a new algorithm to estimate the Weingarten map for point clouds sampled from some distribution on a submanifold embedded in Euclidean space. A statistical model was also established to investigate the optimal convergence rate of the Weingarten map estimator. Numerical experiments were carried out to validate the consistency of the algorithm. Compared with other methods concerning the second fundamental form and curvature estimation, our method showed great robustness and efficiency. Applications to real data sets indicated that our method worked well in practice.

The Weingarten map is a vital tool which is indispensable in the study for submanifolds. In the literature of manifold learning, where data sets from high dimensional Euclidean space are always assumed to be sampled from some underlying low dimensional manifold, the Weingarten map helps researchers to investigate the underlying space from an extrinsic view. Furthermore, the Weingarten map is also in close relation with many other concepts in Riemannian geometry. Curvature is mentioned as one of them. Therefore, we expect many possible applications in the future. For example, the followings are of our concern. 
  
\textbf{Optimization on Manifolds}. In \cite{absil2013extrinsic} the authors presented a useful relationship between the Weingarten map and Riemannian hessian of functions, and the latter plays an important role in Riemannian optimization. Especially, Riemannian hessian is critical in all kinds of second-order optimization methods on Riemannian manifolds \cite{boumal2020introduction,huang2015broyden}. We expect our method can be applied in Riemannian optimization problems such as low-rank matrix completion \cite{vandereycken2013low}, high-dimensional tensor completion \cite{steinlechner2016riemannian} and independent subspace analysis \cite{nishimori2006riemannian}. 

\textbf{Laplacian-based Methods}. Laplacian-based methods are extensively studied in manifold learning. Theories and applications about graph Laplacian are popular in the realm of computer vision \cite{dinesh2020point}, network analysis\cite{grindrod2018deformed}, and statistical learning \cite{belkin2002laplacian,belkin2007convergence,belkin2004semi}. It is known that the Weingarten map is related to the Beltrami-Laplacian operator on Riemannian manifolds. We expect to find other ways to construct the Laplacian estimator on point clouds. 

\textbf{Statistical Inference on Point Clouds}. In the application for brain cortical data our method showed great efficiency in curvature estimation. In recent researches, many kinds of curvature appeared to be useful indexes in statistical inference and regression \cite{luders2006curvature,yue2020curvature}. We expect to find applications in fields such as biostatistics, medical imaging and neuroscience.


 \appendix
 \appendixpage
\renewcommand{\appendixname}{Appendix~\Alph{section}}
\section{The Second Fundamental Form, Weingarten Map and Curvature}\label{appen-weingarten}

Let $\mm\subseteq \mathcal{N}$ be a submanifold with induced Riemannian metric. The connection in $\mathcal{N}$ is denoted by $\overline{\nabla}$ and the Riemannian connection on $\mm$ is denoted by $\nabla$. Let $\mathfrak{X}(\mm)$ be the set of vector fields on $\mm$. For any vector fields $X,Y\in\mathfrak{X}(\mm)$, we have \emph{the Gauss formula}
\begin{equation}\label{gaussformula}
\overline{\nabla}_X Y=\nabla_XY+h(X,Y)
\end{equation}
where
\begin{equation}
\nabla_XY=\left(\overline{\nabla}_XY\right)^\top,h(X,Y)=\left(\overline{\nabla}_XY\right)^\bot
\end{equation}
The map defined by 
\begin{equation}
h:\mathfrak X(\mm)\times \mathfrak X(\mm)\to \Gamma(T^\bot \mm)
\end{equation}
has the following properties:
\begin{equation}
\begin{cases}
h(X+Y,Z)=h(X,Z)+h(Y,Z)\\
h(\lambda X,Y)=\lambda h(X,Y),\forall \lambda\in C^\infty(\mm)\\
h(X,Y)=h(Y,X)
\end{cases}
\end{equation}
Therefore, $h$ is a second-order symmetric covariant tensor field on $\mm$.

\begin{definition}
	The tensor field $h$ is called the second fundamental form of submanifold $\mm$. Especially, at every point $p\in \mm$, the second fundamental form defines a symmetric, bilinear map
	\begin{equation}
	h:T_p\mm\times T_p\mm\to T_p^\bot \mm
	\end{equation} 
\end{definition}

Let $\xi\in\Gamma(T^\bot \mm)$ be a normal vector field. We have \emph{the Weingarten formula}
\begin{equation}\label{weinformula}
\overline{\nabla}_X\xi=-A_\xi(X)+\nabla_X^\bot\xi
\end{equation} 
where
\begin{equation}
A_\xi(X)=\left(\overline{\nabla}_X^\bot\right)^\top,\nabla_X^\bot \xi=\left(\overline{\nabla}_X\xi\right)^\bot
\end{equation}

\begin{theorem}
 For every $\xi\in \Gamma(T^\bot \mm)$, the map defined by 
\begin{equation}
A_\xi:\mathfrak X(\mm)\to \mathfrak X(\mm)
\end{equation}
is a smooth $(1,1)$-form tensor field. At every point $p\in \mm$, there is a self-adjoint transformation
\begin{equation}
A_\xi:T_p\mm\to T_p\mm
\end{equation} 
 and satisfies the identity
\begin{equation}
\langle A_\xi(v),w\rangle=\langle h(v,w),\xi\rangle,\forall v,w\in T_p\mm
\end{equation}

Therefore, at every point $p\in \mm$,  there is a bilinear map $A:T_p^\bot \mm\times T_p \mm\to T_p\mm$ such that
\begin{equation}
(\xi,v)\mapsto A(\xi,v)=A_\xi(v)
\end{equation}	
\end{theorem}
\begin{definition}
	 At $p\in \mm,\xi \in T_p^\bot \mm$, the linear map $A\xi:T_p\mm\to T_p\mm$ is called the shape operator or the Weingarten map at $p$ with respect to $\xi$.
\end{definition}

Let $e_1,\cdots,e_m$ be a basis of the tangent space and $\xi_1,\cdots,\xi_{d-m}$ be a basis of the normal space. Assume
\begin{equation}
	A_{\xi_\alpha}e_i=\sum_{j=1}^mA_{\alpha i}^je_j
\end{equation}
Then
\begin{equation}
	\begin{aligned}
	h(e_i,e_j)=\sum_{\alpha=1}^{d-m}A_{\alpha i}^j\xi_\alpha
	\end{aligned}
\end{equation}
\begin{definition}
	Let
	\begin{equation}
	H=\frac{1}{m}{\rm{tr}}(h)=\frac{1}{m}\sum\limits_i h(e_i,e_i)=\frac{1}{m}\sum\limits_{i,\alpha}h_{ii}^\alpha e_{\alpha}
	\end{equation}
	then $H$ is independent to the choice of orthonormal fields, and is called the mean curvature vector field.
\end{definition}

According to the definition, we have
\begin{equation}
	H=\frac{1}{m}\sum_{i=1}^m h(e_i,e_i)=\frac{1}{m}\sum_{i=1}^m\sum_{\alpha=1}^{d-m}A_{\alpha i}^i\xi_\alpha=\sum_{\alpha=1}^{d-m}(\frac{1}{m}\text{trace}(A_{\xi_\alpha})\xi_{\alpha})=\sum_{\alpha=1}^{d-m}H^\alpha\xi_\alpha
\end{equation}
where $H^\xi=\tilde{g}(H,\xi)$ is called the mean curvature along $\xi$. The value $\|H\|=(\sum_\alpha\|H^\alpha\|^2)^{1/2}=\frac{1}{m}(\sum_{\alpha=1}^{d-m}\text{trace}(A_{\xi_{\alpha}})^2)^{1/2}$ is called the mean curvature. If $M$ is a hypersurface, $\|H\|=\frac{1}{m}|\text{trace}(A_{\xi_{\alpha}})|$. For a surface in 3-dimensional Euclidean space, this coincides with \emph{the absolute mean curvature}. 

Let $R$ and $\overline{R}$ be the Riemann curvature tensor of $\mm$ and $\mathcal{N}$ respectively. From the equations \eqref{gaussformula} and \eqref{weinformula} we can derive \emph{the Gauss equation}

\begin{equation}
	\overline{R}(X,Y,Z,W)=R(X,Y,Z,W)+\langle h(X,Z),h(Y,W)\rangle-\langle h(Y,Z),h(X,W)\rangle
\end{equation}
Especially the sectional curvature can be expressed as 
\begin{equation}
	\overline{K}(X,Y)=K(X,Y)-\langle h(X,X),h(Y,Y)\rangle+\|h(X,Y)\|^2
\end{equation}
If the ambient space is Euclidean, then $\overline{R}$ vanishes identically. Let $e_i,e_j$ span a two-plane $\pi_{ij}$, then the sectional curvature of $\pi_{ij}$ is 
\begin{equation}
	K(\pi_{ij})=-\|h(e_i,e_j)\|^2+\langle h(e_i,e_i),h(e_j,e_j)\rangle=\sum_{\alpha=1}^{d-m}(-(A_{\alpha i}^j)^2+A_{\alpha i}^iA_{\alpha j}^j)
\end{equation}
From $A_{\xi_{\alpha}}$ we extract the $2\times 2$ submatrix $A_{\xi_{\alpha}}|_{\pi_{ij}}$ with $i$th and $j$th row and column. Then 
\begin{equation}
	K(\pi_{ij})=\sum_{\alpha=1}^{d-m}\det(A_\alpha|_{\pi_{ij}})
\end{equation}

\begin{example}[Hypersurfaces]
	If $\mm$ is a hypersurface of $\mathbb{E}^d$ with a unit normal vector field $\xi$, the sectional curvature is $K(\pi_{ij})=\det(A_\xi|_{\pi_{ij}})$. If $e_1,\cdots,e_{d-1}$ is a basis that diagonalizes the Weingarten map with eigenvalues $\lambda_1,\cdots,\lambda_{d-1}$, then $K(\pi_{ij})=\lambda_i\lambda_j$.  The eigenvalues are called \emph{principal curvature} and eigenvectors are called \emph{principal directions}. The determinant of $A_\xi$ is called \emph{the Gauss-Kronecker curvature}.
\end{example}

\begin{example}[Planar Curves]
	A planar curve is a 1-dimensional manifold embedded in the 2-plane. Let $\mathbf{t}$ be the tangent vector field and $\mathbf{n}$ be the normal vector field. We have 
	\begin{equation}
			A_{\mathbf{n}}\mathbf{t}=-(\overline{\nabla}_{\mathbf{n}}\mathbf{t})^\top=\kappa \mathbf{t}
	\end{equation}
	where $\kappa$ is the curve curvature. Thus the estimation can be simply given by 
	\begin{equation}
			\hat{A}=\frac{\Delta\mathbf{n}\cdot\mathbf{t}}{\Delta p\cdot\mathbf{t}}\approx \kappa
	\end{equation}
\end{example}

\begin{example}[Space Curves]
	A space curve is a 1-dimensional manifold embedded in 3-space. Let $\mathbf{t}$ be the tangent vector field, $\mathbf{n}$ be the normal vector field, and $\mathbf{b}$ be the binormal vector field. We have the Frenet formula
	\begin{equation}
			\frac{d}{ds}\left(\begin{array}{c}\mathbf{t}\\ \mathbf{n}\\ \mathbf{b}\end{array}\right)=\left(\begin{array}{ccc}0&\kappa&0\\ -\kappa&0&\tau\\ 0&-\tau&0\end{array}\right)\left(\begin{array}{c}\mathbf{t}\\ \mathbf{n}\\ \mathbf{b}\end{array}\right)
	\end{equation}
	where $\kappa$ is curvature and $\tau$ is torsion. Let $\xi=\cos(\theta)\mathbf{n}+\sin(\theta)\mathbf{b}$ be a unit normal vector field. We have
	\begin{equation}
			A_\xi\mathbf{t}=\cos(\theta)A_\mathbf{n}\mathbf{t}+\sin(\theta)A_\mathbf{b}\mathbf{t}=\cos(\theta)\kappa\mathbf{t}
	\end{equation}
	Similarly, let $\xi_\perp=-\sin(\theta)\mathbf{n}+\cos(\theta)\mathbf{b}$ be the unit normal vector perpendicular to $\xi$. Then
	\begin{equation}
			A_{\xi_\perp}\mathbf{t}=-\sin(\theta)\kappa\mathbf{t}
	\end{equation}
	By definition the mean vector field is $H=\cos(\theta)\kappa\xi-\sin(\theta)\kappa\xi_\perp=\kappa\mathbf{n}$, and the mean curvature is $\|H\|=\kappa$. As a byproduct we obtain the normal vector $\mathbf{n}=H/\|H\|$.
\end{example}

\section{The Exponential Map}\label{appendix-exponetial}
 
The second fundamental form is closely related to the injective radius $\iota(\mm)$, as proved in \cite{alexander2006gauss}: If the operator norm $\|h\|\le C$, then $\iota(\mm)\ge\pi/C$. Thus the exponential map can be defined on the ball of radius $\pi/C$ in $T_p\mathcal{M}$ for all $p$.

The exponential map for submanifolds embedded in Euclidean spaces can be expressed in special forms. Fix $p\in\mathcal{M}$. Without loss of generality we transform $\mathcal{M}$ isometrically in $\mathbb{R}^d$ so that $p$ is the origin and the first $m$ standard basis $\{e_1,\cdots,e_m\}$ spans the tangent space $T_p\mathcal{M}$. Consider the coordinate components of the exponential map
\begin{equation}
	\br(u^1,\cdots,u^m)=(r^1(u^1,\cdots,u^m),\cdots,r^d(u^1,\cdots,u^m))
\end{equation}
We can express the derivatives of component functions using the geometric terms defined on $\mathcal{M}$. In fact, we have the following proposition.

\begin{prop}\label{exponent-map-prop}
	Let $\bu=(u^1,\cdots, u^m,0,\cdots,0)\in T_p\mm\subseteq \mathbb{R}^d$. The exponential map for $\mm$ can be expressed by
	\begin{equation}
		\br(\bu)=p+\bu+\frac{1}{2}h(\bu,\bu)-\frac{1}{6}A_{h(\bu,\bu)}(\bu)+\frac{1}{6}(\nabla_\bu h)(\bu,\bu)+O(\|\bu\|^4)
	\end{equation}
	Furthermore, the derivatives of component functions can be expressed by
	\begin{equation}
		\begin{aligned}
		& \frac{\partial r^j}{\partial u^i}(0)=\delta^j_i,\quad \frac{\partial r^\alpha}{\partial u^i}(0)=0 \\
		& \frac{\partial^2 r^k}{\partial u^i\partial u^j}(0)=0,\quad \frac{\partial^2 r^\alpha}{\partial u^i\partial u^j}(0)=\langle A_{e_\alpha}e_i,e_j\rangle \\
		&\frac{\partial^3 r^l}{\partial u^i\partial u^j\partial u^k}(0)=-\langle A_{h(e_i,e_j)}(e_k),e_l\rangle,\quad \frac{\partial^3 r^\alpha}{\partial u^i\partial u^j\partial u^k}(0)=\langle(\nabla_{e_i}h)(e_j,e_k),e_l\rangle
		\end{aligned}
	\end{equation}
	for $i,j,k,l=1,\cdots,m;\alpha=m+1,\cdots,n$.
\end{prop}

\begin{proof}
	Firstly, since the tangent map of $\br$ at origin is the identity, we have 
	\begin{equation}
		d\br_0(e_i)=(\partial_ir^1(0),\cdots,\partial_ir^n(0))=e_i
	\end{equation}
	We obtain that $\partial_ir^j(0)=\delta^j_i$ for $i,j=1,\cdots,m$ and $\partial_ir^\alpha(0)=0$ for $\alpha=m+1,\cdots,n$. 
	
	For the second derivatives, let $\bu=\sum_{i=1}^m u^ie_i$ and $\gamma(s)=\br(s\bu)$. we have
	\begin{equation}
			\ddot{\gamma}(0)=(\sum_{i,j}\partial_{i}\partial_jr^1(0)u^iu^j,\cdots,\sum_{i,j}\partial_i\partial_jr^n(0)u^iu^j)=h(\bu,\bu)
	\end{equation}
	Thus, $\partial_i\partial_jr^k(0)=0$ for $i,j,k=1,\cdots,m$. Let $h_\alpha(\bu,\bv)=\langle h(\bu,\bv),e_\alpha\rangle=\langle A_{e_\alpha}(\bu),\bv\rangle$. We see that $\big[\partial_i\partial_jr^\alpha(0)=\langle A_{e_\alpha}e_i,e_j\rangle\big]_{i,j}$ is the matrix representation for the Weingarten map $A_{e_\alpha}$ for $\alpha=m+1,\cdots,n$. 
	
	By \eqref{weinformula}, we have
	\begin{equation}
			\dddot{\gamma}=-A_{h(\dot{\gamma},\dot{\gamma})}(\dot{\gamma})+\nabla_{\dot{\gamma}}^\perp h(\dot{\gamma},\dot{\gamma})
	\end{equation}
	If we define $\nabla h(\bu,\bv,\bw)=\nabla_\bw^\perp h(\bu,\bv)-h(\nabla_\bw\bu,\bv)-h(\bu,\nabla_\bw\bv)$, then we have
	\begin{equation}
			\dddot{\gamma}(0)=-A_{h(\bu,\bu)}(\bu)+(\nabla h)(\bu,\bu,\bu)
	\end{equation}
	Comparing the components on both sides we obtain that
	\begin{equation}
		\begin{aligned}
			&\partial_i\partial_j\partial_kr^l(0)=-\langle A_{h(e_i,e_j)}(e_k),e_l\rangle\\
		&\partial_i\partial_j\partial_kr^\alpha(0)=\langle(\nabla_{e_i}h)(e_j,e_k),e_l\rangle
		\end{aligned}
	\end{equation}
    for $i,j,k,l=1,\cdots,m$ and $\alpha=m+1,\cdots,n$. We have deduced the Taylor expansion for the exponential map.
\end{proof}

Let $q=\br(s\bu)$ where $\|\bu\|=1$. We can compare the Euclidean distance $\|q-p\|$ and the Riemannian distance $s=d_\mathcal{M}(p,q)$. In fact, from the Taylor expansion of the exponential map, we have
\begin{equation}
	\|q-p\|^2=s^2+\frac{1}{4}\|h(\bu,\bu)\|^2s^4-\frac{1}{3}\langle A_{h(\bu,\bu)}(\bu),\bu\rangle s^4+o(s^4)
\end{equation}
Therefore, we have
\begin{equation}
	\|q-p\|=s-\frac{1}{24}\|h(\bu,\bu)\|^2s^3+o(s^3)
\end{equation}

\bibliographystyle{plain}
\bibliography{example_paper}

\end{document}